\pdfoutput=1
\documentclass[11pt]{article}

\usepackage[]{ACL2023}

\usepackage[normalem]{ulem}
\useunder{\uline}{\ul}{}

\usepackage{times}
\usepackage{latexsym}
\usepackage{todonotes}
\usepackage{subcaption}
\usepackage{tcolorbox}
\usepackage{amsmath}
\usepackage{siunitx}
\usepackage{booktabs}
\usepackage{enumitem}
\usepackage{xspace}
\newcommand{\method}[0]{\textsc{Scaffold}\xspace}

\usepackage{multirow}
\usepackage[T1]{fontenc}
\usepackage[utf8]{inputenc}

\usepackage{microtype}

\usepackage{inconsolata}

\title{Scaffolding Coordinates to Promote Vision-Language Coordination \\ in Large Multi-Modal Models}

\author{
Xuanyu Lei$^{1,2}$, 
Zonghan Yang$^{1}$, 
Xinrui Chen$^{1}$, 
Peng Li$^{2}$\footnotemark[1],
Yang Liu$^{1,2}$\footnotemark[1] \\ \\
$^1$Department of Computer Science and Technology, Tsinghua University, Beijing, China \\
$^2$Institute for AI Industry Research (AIR), Tsinghua University, Beijing, China
}

\begin{document}
\maketitle

\footnotetext[1]{Corresponding author: lipeng@air.tsinghua.edu.cn}
\footnotetext[2]{Corresponding author: liuyang2011@tsinghua.edu.cn}

\begin{abstract}
State-of-the-art Large Multi-Modal Models (LMMs) have demonstrated exceptional capabilities in vision-language tasks. Despite their advanced functionalities, the performances of LMMs are still limited in challenging scenarios that require complex reasoning with multiple levels of visual information. Existing prompting techniques for LMMs focus on either improving textual reasoning or leveraging tools for image preprocessing, lacking a simple and general visual prompting scheme to promote vision-language coordination in LMMs. In this work, we propose \method prompting that scaffolds coordinates to promote vision-language coordination. Specifically, \method overlays a dot matrix within the image as visual information anchors and leverages multi-dimensional coordinates as textual positional references. Extensive experiments on a wide range of challenging vision-language tasks demonstrate the superiority of \method over GPT-4V with the textual CoT prompting. Our code is released in https://github.com/leixy20/Scaffold.
\end{abstract}

\section{Introduction}

Large Multi-Modal Models (LMMs) like GPT-4V~\cite{openai2023gpt4} and Gemini~\cite{geminiteam2023gemini} have demonstrated impressive zero-shot capabilities in processing diverse visual-language tasks. Leveraging the advanced reasoning ability of the language model component, early attempts have been made to deploy LMMs in realistic scenarios, such as autonomous driving~\cite{wen2023road} and anomaly detection~\cite{cao2023generic}.

\begin{figure}[t]
    \centering
    \includegraphics[width=\linewidth]{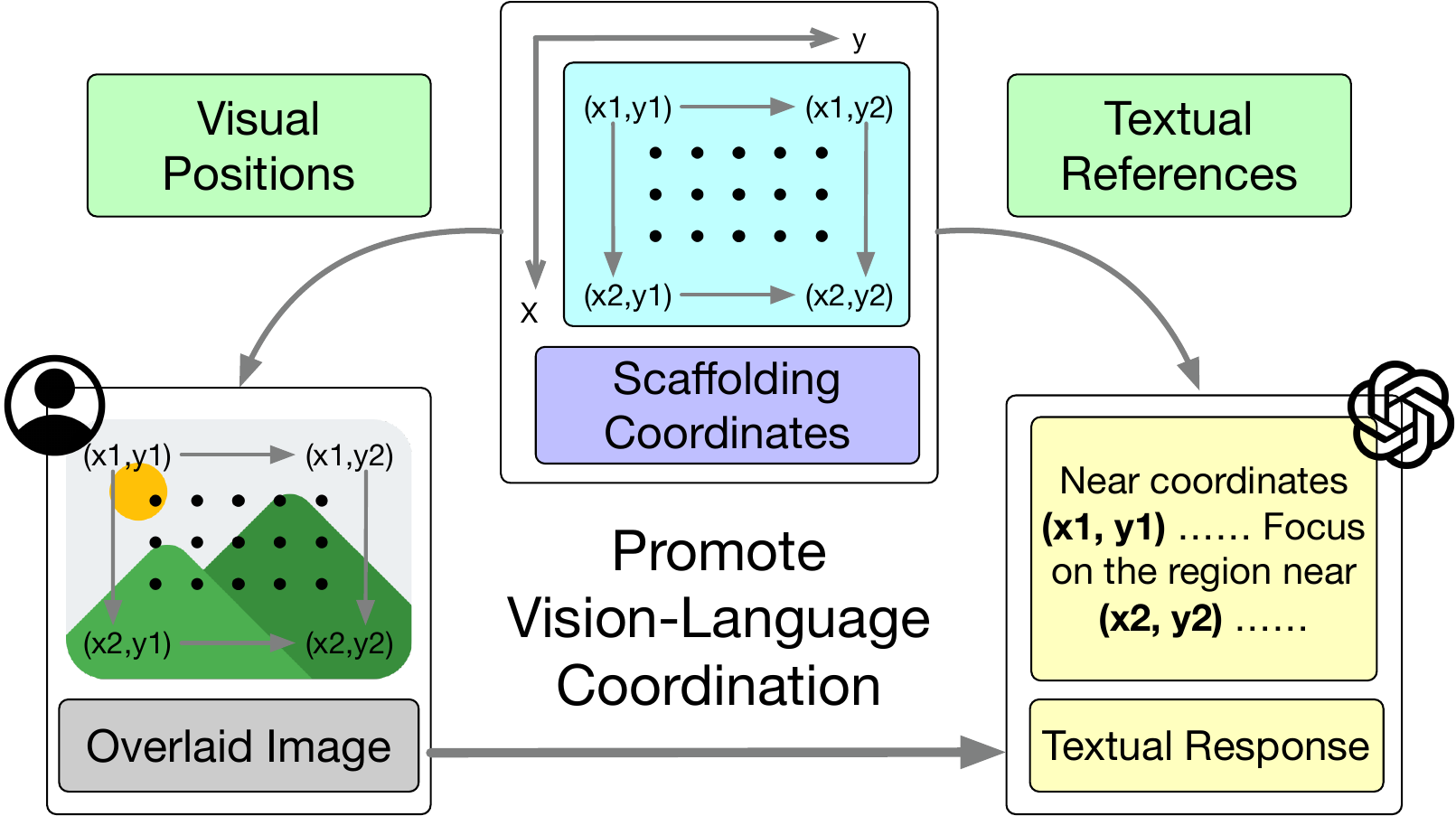}
    \caption{The overall framework of \method. \method overlays a dot matrix onto the input image with Cartesian coordinates labeled aside the dots. The coordinates are also briefed in the textual prompt, which steers the LMM to leverage the dots on the image as a scaffold and promotes vision-language coordination.}
    \label{fig:overall}
    \vspace{-5mm}
\end{figure}

\definecolor{darkgreen}{rgb}{0.0, 0.5, 0.0}
\begin{figure*}[ht]
\centering
\begin{tcolorbox}
    \newlength{\imageheight}
    \settoheight{\imageheight}
    {\includegraphics[width=.5\linewidth]{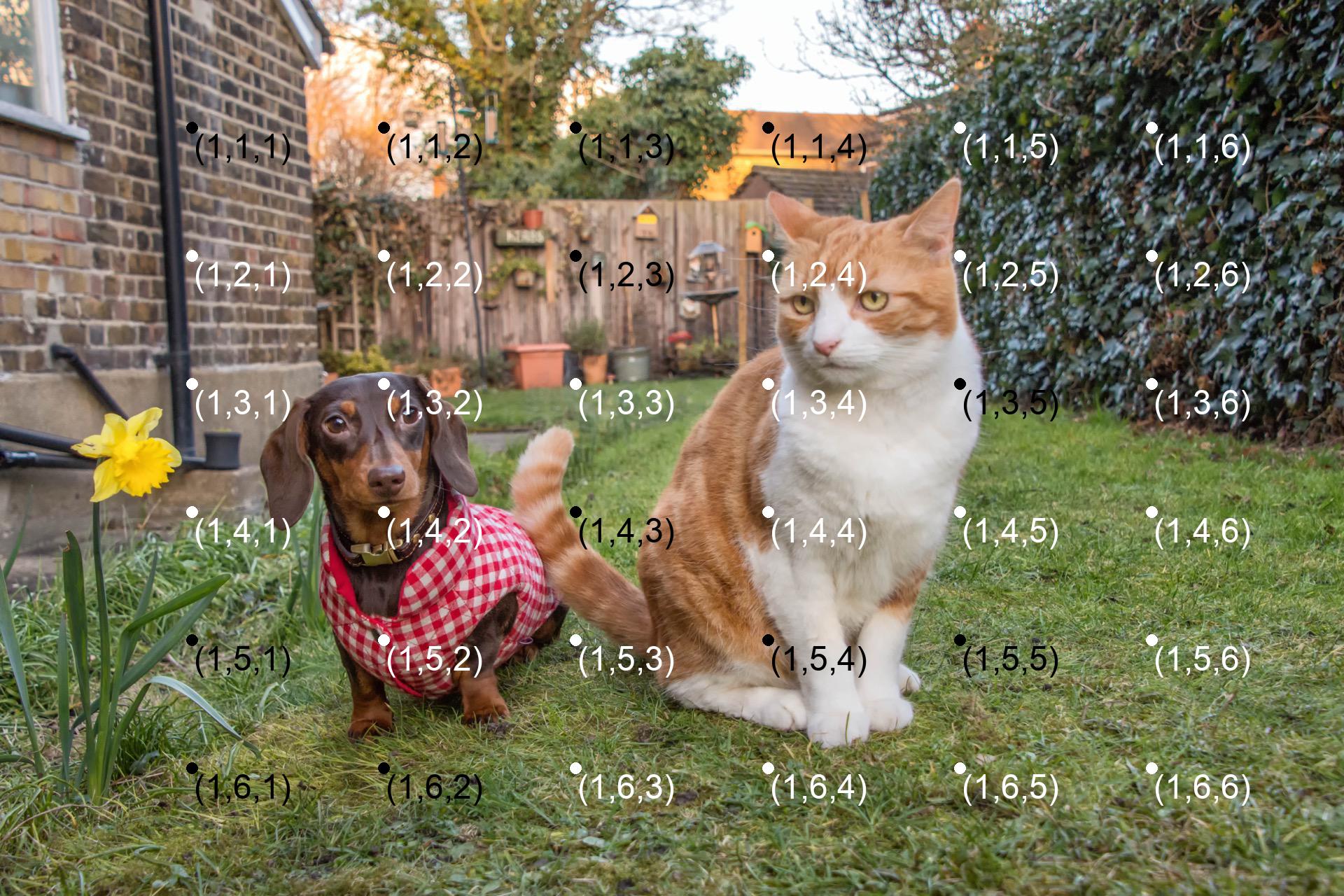}}
    
    \includegraphics[height=\imageheight]{figs/wino_img_0_dots.jpg}
    \hfill % This will insert a space between the two images
    \includegraphics[height=\imageheight]{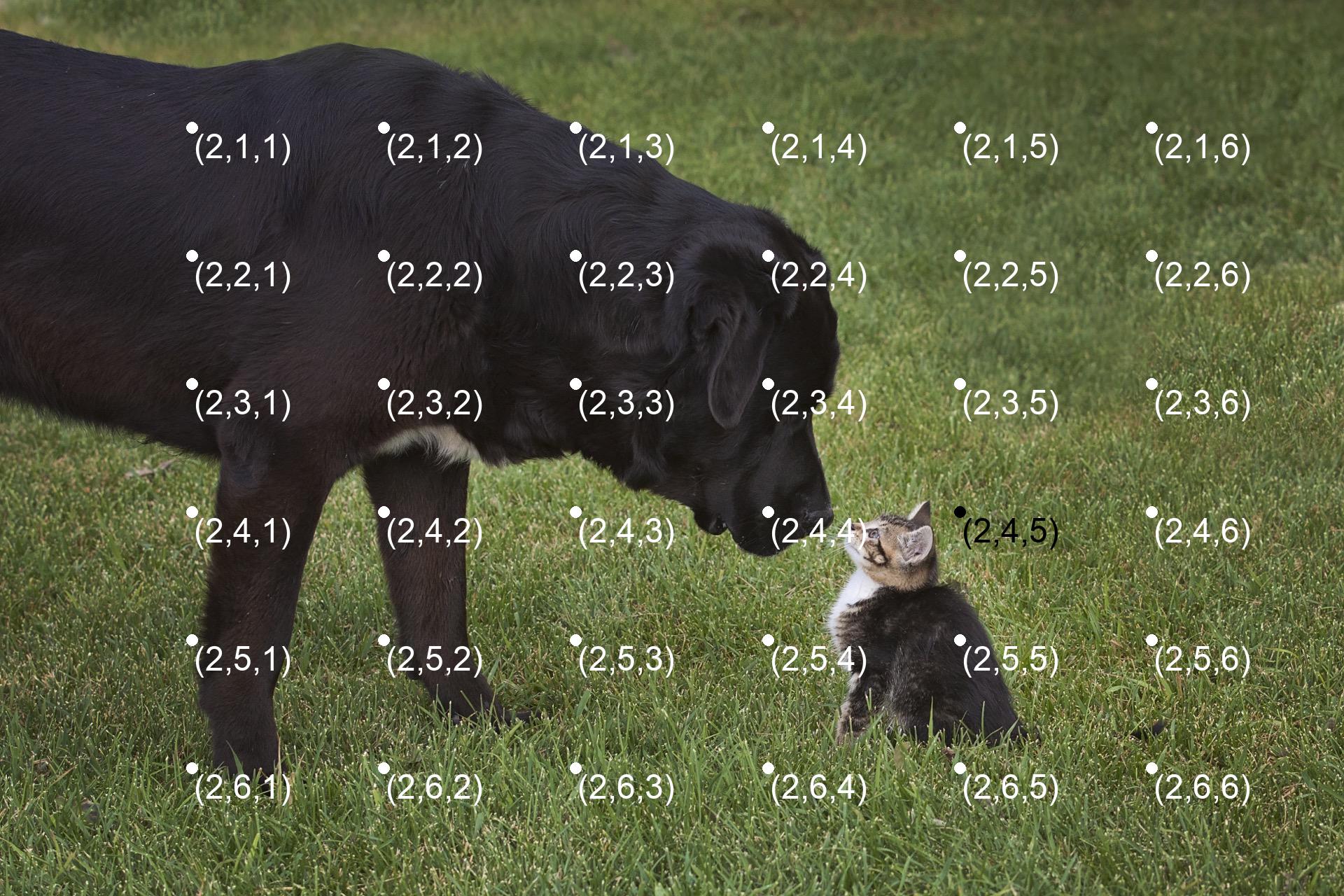}
    %\includegraphics[width=0.46\textwidth]{figs/wino_img_0_dots.jpg}
    %\hfill % This will insert a space between the two images
    %\includegraphics[width=0.46\textwidth]{figs/wino_img_1_dots.jpg}
    \small
    {\textit{\textbf{Question}: Choose the correct image for the caption. Caption: a big cat is next to a small dog. Options: (A) image 1(left) (B) image 2(right) }}

    \vspace{2mm}
    \small
    \textbf{GPT-4V + \method}:
    \vspace{1mm}
    
    \textbf{Textual Prompt for \method}:
    Two images are provided, each overlaid with a grid of dots arranged in a matrix with dimensions 6 by 6. Each dot on this grid is assigned a unique set of three-dimensional coordinates labeled as (t, x, y). ...
    
    1. When you mention any key objects in the image, first output their nearest coordinates then identify them.

    2. You can use the coordinates to determine the spatial relationships of the objects. ...
    
    \vspace{1mm}
    \textbf{Output}: In the first image (1), \textcolor{darkgreen}{at coordinates (1,4,2) and (1,5,4), there is a small dog wearing a red and white checkered outfit and a large orange and white cat sitting next to each other on the grass.} This matches the caption "a big cat is next to a small dog.". In the second image (2), \textcolor{darkgreen}{at coordinates (2,3,3) and (2,5,5), there is a large black dog and a small kitten facing each other on the grass}. Therefore, \textcolor{darkgreen}{the correct image for the caption is: A.}

    \vspace{-1mm}
    \tcblower
    \vspace{-1mm}

    \small
    \textbf{GPT-4V Output}: \textcolor{red}{The correct image is: (B) image 2(right).} This image shows a large black dog next to a much smaller kitten on a grassy surface. The size difference between the two animals is evident, \textcolor{red}{with the dog being the "big cat" and the kitten being the "small dog" in a metaphorical sense}, as the caption likely plays on the term "big cat" which usually refers to large wild felines. \textcolor{red}{The first image (A) shows an actual cat and a small dog, which does not fit the caption.}% as described.}

\end{tcolorbox}
\vspace{-0.2cm}
\caption{An example of \method in double-image vision-language tasks, with correct textual reasoning illustrated in \textcolor{darkgreen}{green} and incorrect textual reasoning illustrated in \textcolor{red}{red}. Note that only the original images and questions are visible to \textbf{GPT-4V}; \textbf{\method} adds the coordinates on images and the corresponding textual prompt guidance.}
\vspace{-0.35cm}
\label{fig:main_figure}
\end{figure*}

However, current LMMs display limited performance when conducting complex reasoning over multiple levels of visual information~\cite{yang2023dawn, wu2023early, wu2023v}. For example, in a spatial reasoning task~\cite{liu2023vsr}, an LMM needs to verify or generate the statement by elucidating the relationship between different sources of visual information, and aligning its internal workings with textual expressions. Challenges for LMMs arise in orchestrating precise visual perception with accurate language understanding and generation.

\looseness=-1 To enhance vision-language coordination, prior efforts for LMMs can be divided into two categories: \textit{instruction tuning} and \textit{prompting}. \textit{Instruction tuning} uses high-quality image-text pairs of either general purposes~\cite{dai2023instructblip,liu2023llava,liu2023improvedllava,multiinstruct} or specialized domains~\cite{gpt4roi,chen2023positionenhanced,chen2024spatialvlm} to facilitate additional training of LMMs for improved performance. Nevertheless, \textit{instruction-tuning} consumes heavy loads of computing resources and thus sacrifices flexibility in methodology. \textit{Prompting} for LMMs, on the contrary, steers the improved functioning of LMMs in a non-parametric manner. While techniques for language model prompting like Chain-of-Thought~\cite{wei2022chain} apply to LMMs as well, the elicited intermediate reasoning steps mainly take place with the condition of the textual prompts~\cite{wu2023role}. As textual prompting techniques are actively being exploited, few endeavors have been made into visual prompting, which steers the precise visual perception of LMMs for vision-language coordination.

The challenge of visual prompting for LMMs lies in the mismatch of semantic granularity between visual and textual information. While each word is explicitly separated in a textual sentence, different identities in an image are not isolated with clear boundaries. Recent works on visual prompting include leveraging tools to narrow the semantic granularity gap between visual and textual inputs. \citet{yang2023set} leverage advanced image segmentation models~\cite{kirillov2023segment} to construct object segmentation overlays on the input image. \citet{mitra2023compositional} treat the LMM itself as a scene graph extractor to generate visual information in the textual format. However, tool usage inevitably results in additional resource burdens and potentially erroneous information. An alternative avenue of recent efforts is visual search~\cite{wu2023v,google2024pivot}, where the solution to a complex visual task is cast as an iterative search process that accounts for multiple aspects of the image. Nevertheless, the iterative queries of LMMs throughout the visual search process entail considerable expenses, limiting the practical value. Therefore, it remains elusive whether a simple and general visual prompting scheme exists to promote vision-language coordination in LMMs.

In this work, we present \method, a simple and versatile visual prompting scheme to promote the coordination between vision and language in LMMs. \method overlays a dot matrix onto the input image, and each dot is labeled with its multi-dimensional Cartesian coordinate. The dot matrix on the image forms the scaffold that indicates relative visual positions for LMMs. The overlaid coordinates are also included in the textual prompt, which explicitly strengthens the connection between visual and textual information for LMMs. The LMMs are thus steered to leverage the coordinates to solve different vision-language tasks. In this way, \method provides a scaffold to promote vision-language coordination in LMMs. Extensive experiments on spatial reasoning, compositional reasoning, fine-grained grounding, and hallucination benchmarks demonstrate the superiority of \method over GPT-4V with the textual CoT prompting. We also show that the performance of \method can be further enhanced with region cropping, which reveals the promising future of active perception enabled by \method.

\section{Related Work}
\textbf{Large Multi-Modal Models (LMMs).}
State-of-the-art LMMs like GPT-4V~\cite{openai2023gpt4} and Gemini~\cite{geminiteam2023gemini} have excelled in general vision-language tasks~\cite{wu2023early, yang2023dawn, fu2023challenger}. The integration of visual capabilities in LMMs with advanced language proficiency and instruction-following skills pave the way for versatile visual interactive agents, both in digital~\cite{he2024webvoyager, zheng2024gpt4vision} and embodied environments~\cite{wake2023gpt4vision, chen2023endtoend}.

\noindent\textbf{GPT-4V Evaluation.}
As a leading LMM, GPT-4V~\cite{openai2023gpt4} has significantly expanded the boundaries of LMM capabilities, motivating researchers to systematically explore its strengths and weaknesses~\cite{wu2023early, yang2023dawn}. Despite its proficiency, researchers have proposed challenging benchmarks that reveal large performance gap between GPT-4V and humans, including MMVP~\cite{tong2024eyes}, MMMU~\cite{yue2023mmmu}, Mementos~\cite{wang2024mementos}, V* Bench~\cite{wu2023v}, Contextual~\cite{wadhawan2024contextual}, etc. Extensive evaluations indicate that plenty of room exists for state-of-the-art LMMs to improve their certain visual-language capabilities. Additionally, previous works~\cite{yan2023multimodal, liu2023holistic, cao2023generic, wen2023road} have shown a promising future of potential applications of GPT-4V in downstream domains like medicine science.

\noindent\textbf{Multi-Modal Prompting Methods.}
Prompting methods focus on unlocking model potentials by carefully constructing model inputs. Chain-of-Thought~\cite[CoT]{wei2022chain, kojima2022large} and its variants~\cite{yao2023tree, besta2023graph} have successfully elicited reasoning capabilities in language models. However, in multi-modal contexts such as compositional reasoning, the original CoT is less effective~\cite{mitra2023compositional}. Consequently, numerous multi-modal prompting methods have been developed for specific visual capabilities. For instance, Compositional CoT~\cite{mitra2023compositional} for compositional reasoning, Spatial CoT~\cite{chen2024spatialvlm} for spatial understanding, Set-of-Marks prompting~\cite{yang2023set} for visual grounding. However, these methods tend to be tailored for specific capabilities, calling for simple and general visual prompting schemes.

\section{Methodology}
In this section, we introduce \method prompting for vision-language coordination in LMMs. 

\subsection{Visual Perspective of \method: \\\ \ \ \ \ \ \ \ \ Dot Matrices and Coordinates}

Visually, we enhance each input image with a uniformly distributed rectangular dot matrix, where each dot is labeled with multi-dimensional coordinates. These dots serve as visual positional anchors, while their coordinates are utilized as textual references in textual responses. 

\noindent\textbf{Visual Anchor Implementation.}
We select rectangular dot matrices as our visual anchor due to their simplicity, flexibility in textual description, and potential adaptability to image sequences. Unlike grids, which divide images into separate regions and may disrupt continuous visual content, dot matrices offer a less intrusive overlay. Additionally, to preserve original information, we deliver both the original image and the coordinates-overlaid image as inputs in single-image tasks.

\noindent\textbf{Coordinates Implementation.}
For our approach, we use multi-dimensional Cartesian coordinates due to its simplicity and clarity. 
For a single image with an overlaid dot matrix of size $h\times w$, we assign two-dimensional coordinates $(x,y)$ to each dot, representing its relative visual position. Here, the x-coordinate ascends from $1$ to $h$ within each column, while the y-coordinate ascends from $1$ to $w$ within each row.
For image sequences, we extend these coordinates to three-dimensional $(t,x,y)$. The t-coordinate remains constant within each image but increases sequentially across the sequence, allowing for differentiation between images and enhancing temporal perception.

In comparison, we also consider other coordinate options and identify their limitations. Absolute pixel coordinates, for example, consume excessive space and are complicated to perceive and apply accurately. Furthermore, one-dimensional Cartesian coordinates and alphabetic coordinates fall short in providing adequate positional information.

\noindent\textbf{Other Factors.} 
\textit{1. Matrix Size.} The matrix should be visually clear and provide ample space for multi-dimensional coordinates. For simplicity, we employ a $6 \times 6$ matrix for general vision-language tasks. 
\textit{2. Matrix Density.} Without prior visual knowledge, we choose rectangular dot matrices with a uniform density for general vision-language tasks, providing LMMs equal assistance when reasoning across different regions. 
\textit{3. Matrix Color.} The coordinates are designed to be recognizable by LMMs using their OCR capabilities. Consequently, we color each dot in either black or white according to its contrast against the background.

\subsection{Textual Perspective of \method: Description and Guidelines}
To complement the coordinates-overlaid visual inputs, we prepend textual guidance to task instructions for LMMs. This includes a brief description of the dot matrices and coordinates, accompanied by several general guidelines for their effective use, as detailed in Appendix~\ref{appendix:scaffold_prompts}. The characteristics of these descriptions and guidelines are as follows: \textbf{(1) Conciseness:} The textual guidance is deliberately brief and clear, ensuring easy comprehension. \textbf{(2) Generality:} Designed to be universally applicable, these guidelines are not specific to any particular scenario, making them suitable for a wide range of vision-language tasks. \textbf{(3) Extensibility:} The guidelines are semantically independent, allowing for the seamless addition of more tailored instructions based on different scenarios. \textbf{(4) Compositionality:} The prepended texts can be easily combined with other prompting methods, such as zero-shot or compositional CoT~\cite{kojima2022large,mitra2023compositional}.

\section{Experiments}

\setlist[itemize]{itemsep=0em, topsep=0.2em}

\definecolor{deepgreen}{rgb}{0.0, 0.5, 0.0}
\begin{table*}[t]
\centering\small
% \resizebox{\linewidth}{!}{
\begin{tabular}{c|ccc|ccc}
\toprule
\textbf{Crucial Capability}                                                                 & \textbf{Dataset}                     & \textbf{Size} & \textbf{Metric}  & \textbf{Naive} & \textbf{CoT}  & \textbf{\method (Ours)} \\ \midrule
                                                                                         & MME (Position)                        & 60            & Accuracy         & 51.7           & 51.7          & \textbf{75.0} \textcolor{deepgreen}{\scriptsize (+23.3)}      \\
                                                                                         & VSR                                  & 200           & Accuracy         & 67.8           & 70.4          & \textbf{74.4} \ \textcolor{deepgreen}{\scriptsize (+6.6)}    \\
\multirow{-3}{*}{\begin{tabular}[c]{@{}c@{}}Spatial \\ Reasoning\end{tabular}}           & EgoThink (Spatial)                    & 50            & LLM as Judge     & 66.0           & 74.0          & \textbf{76.0} \textcolor{deepgreen}{\scriptsize (+10.0)}     \\ \midrule
                                                                                         & Winoground                           & 100           & Group Score      & 17.0           & \textbf{33.0}          & \textbf{33.0} \textcolor{deepgreen}{\scriptsize (+16.0)} \\
                                                                                         & WHOOPS! VQA                         & 200           & BEM Score        & 58.6           & 57.6          & \textbf{62.7} \ \textcolor{deepgreen}{\scriptsize (+4.1)}      \\
\multirow{-3}{*}{\begin{tabular}[c]{@{}c@{}}Compositional \\ Reasoning\end{tabular}}     & CLEVR                                & 200           & LLM as Judge     & 43.5           & 43.0          & \textbf{48.0} \ \textcolor{deepgreen}{\scriptsize (+4.5)}     \\ \midrule
                                                                                         & V* Bench     & 238           & Accuracy         & 27.2           & 30.8        & \textbf{44.6} \textcolor{deepgreen}{\scriptsize (+17.4)}     \\
\multirow{-2}{*}{\begin{tabular}[c]{@{}c@{}}Fine-Grained \\ Visual Understanding\end{tabular}} & Spotting Differences                 & 100           & Accuracy         & 13.0           & 14.0          & \textbf{19.0} \ \textcolor{deepgreen}{\scriptsize (+6.0)}     \\ \midrule
                                                                                         & POPE (Adversarial)                    & 100           & Accuracy         & 79.0           & 80.0          & \textbf{86.0} \ \textcolor{deepgreen}{\scriptsize (+7.0)}      \\
                                                                                         & HallusionBench (Hard)                       & 504           & Accuracy         & 45.6           & 48.8          & \textbf{53.0} \ \textcolor{deepgreen}{\scriptsize (+7.4)}      \\
\multirow{-3}{*}{Hallucination}                                                          & Mementos                             & 100           & LLM as Judge       & 33.3           & 33.5          & \textbf{36.1} \ \textcolor{deepgreen}{\scriptsize (+2.8)}     \\ \midrule
\textbf{Overall}                                                                         & \textbf{All} & \textbf{1852} & \textbf{Average} & \textbf{45.7}  & \textbf{48.8} & \textbf{55.3} \ \textcolor{deepgreen}{\scriptsize (+9.6)}     \\ \bottomrule
\end{tabular}
\vspace{-2mm}
% }
\caption{Results of \method on 11 challenging vision-language benchmarks, with the highest score \textbf{bold}.}
% \vspace{-5mm}
\label{tab:results}
\end{table*}

To demonstrate the effectiveness of \method, we conduct extensive experiments on top of GPT-4V on a range of challenging vision-language tasks, including \textit{Spatial Reasoning}, \textit{Compositional Reasoning}, \textit{Fine-Grained Visual Understanding} and \textit{Hallucination}.  Specifically, we perform systematic evaluations on 11 benchmarks, with detailed information presented in Appendix~\ref{appendix:datasets}. We set the temperature of GPT-4V as zero in our experiments.

\subsection{Benchmarks}
This subsection briefly introduces the benchmarks used for evaluation. Due to the limited budget, for some datasets, we sample a subset for experiments. 

\noindent\textbf{Spatial Reasoning} evaluates LMM capability to infer spatial relationships between objects. Selected benchmarks are as follows. \textit{1. MME (Position split)}~\cite{fu2023mme} is a subset of the MME comprehensive evaluation suite for LMMs to infer object positions.
\textit{2. Visual Spatial Reasoning (VSR)}~\cite{liu2023vsr}  challenges LMMs with 66 types of spatial relations to verify spatial propositions.
\textit{3. EgoThink (Spatial split)}~\cite{cheng2023visionlanguage} tests the spatial reasoning ability of LMMs from a first-person perspective.

\noindent\textbf{Compositional Reasoning} requires LMMs to identify object attributes and their interrelations. Selected benchmarks are as follows.
\textit{1. Winoground~\cite{thrush2022winoground}} is a challenging benchmark that necessitates compositional reasoning of LMMs to match images with captions, reformulated as binary choice questions for our evaluation.
\textit{2. WHOOPS! VQA}~\cite{bittonguetta2023breaking} involves compositional reasoning over commonsense-defying images.
\textit{3. CLEVR}~\cite{johnson2016clevr} is designed for assessing compositional reasoning in program-generated scenes.

\noindent\textbf{Fine-Grained Visual Understanding} requires LMMs to perform visual search and precisely perceive fine-grained visual details. Selected benchmarks are as follows. \textit{1. V* Bench}~\cite{wu2023v} requires LMMs to identify and reason with fine-grained visual details in high-resolution images. \textit{2. Spotting Differences~\footnote{https://www.crazygames.com/game/find-the-difference}} is our newly-collected dataset challenging LMMs to find and pinpoint differences between two similar images, with further details in Appendix~\ref{appendix:spotting_differences}.

\noindent\textbf{Hallucination} measures the tendency of LMMs to generate hallucinatory or illusory perceptions. Selected benchmarks are as follows.
\textit{1. POPE (Adversarial Subset)~\cite{li2023evaluating}} assesses object hallucination by querying the existence of specific objects.
\textit{2. HallusionBench~\cite{guan2023hallusionbench}} consists of meticulously crafted images to measure hallucination and visual illusion in LMMs.
\textit{3. Mementos~\cite{wang2024mementos}} evaluates LMM to conduct precise reasoning over image sequences and measures their performances in terms of object and behavior hallucinations.

\subsection{Baselines}
This section presents the prompting methods utilized as baselines in our experiments, including \textbf{1. Naive Prompting} utilizes original images and user instructions as inputs for LMMs, establishing a straightforward baseline without any prompt optimization. \textbf{2. CoT}~\cite{wei2022chain} guides LMMs to perform step-by-step reasoning before outputting the final answer. The prompt text ``\textit{Let's think step by step.}" is prepended to task descriptions.

\subsection{Results and Analyses}

As presented in Table~\ref{tab:results}, the results demonstrate that \method significantly enhances the visual capabilities of LMMs, surpassing CoT~\cite{wei2022chain} in most evaluated benchmarks. With naive prompting and CoT prompting averaging $45.7$ and $48.8$ respectively, \method successfully obtains an overall improvement of $9.6$.

\definecolor{darkgreen}{rgb}{0.0, 0.5, 0.0}
\begin{figure}[t] % The placement specifier [ht] can be adjusted as needed
\centering
\begin{tcolorbox}
    \small 
    \includegraphics[width=\linewidth]{figs/vsr_example_dots.jpg} 
    \small
    \textit{\textbf{Question}: Determine whether the following statement is true or false: The bench is at the left side of the potted plant.}

    \vspace{-1mm}
    \tcblower
    \small
    
    \textbf{Ground-Truth: } \textcolor{darkgreen}{True.}

    \textbf{GPT-4V + CoT: } The statement is false. In the image provided, \textcolor{red}{the bench is on the right side of the potted plant, not the left}. Therefore, the statement is \textcolor{red}{false}.

    \textbf{GPT-4V + \method: } \textcolor{darkgreen}{The bench is located near coordinates (3,2) through (6,4), and the potted plant is near coordinates (3,6) through (6,6)}. Given the arrangement of the coordinates, the bench is to the left of the potted plant. \textcolor{darkgreen}{true}.
\end{tcolorbox}
\vspace{-0.4cm}
\caption{An example from the Visual Spatial Reasoning (VSR) dataset. \method facilitates better vision-language coordination than other methods.} 
\label{fig:vsr_example}
\vspace{-0.5cm}
\end{figure}

\noindent\textbf{Spatial Reasoning:} \method notably enhances the spatial reasoning capabilities of LLMs across the three benchmarks, with an average improvement of $13.3$. Fig.~\ref{fig:vsr_example} illustrates how \method enabled GPT-4V adeptly identifies crucial objects and records accurate positional information using two-dimensional coordinates, leading to the correct assessment of spatial relations through the numerical analysis of their x-coordinates.

\noindent\textbf{Compositional Reasoning:} With \method, GPT-4V demonstrates improved abilities in compositional reasoning with an average improvement of $8.2$, showing enhanced perception of key visual elements and smoother reasoning processes. As Fig.~\ref{fig:wino_example} in Appendix~\ref{appendix:cases} shows, \method associates crucial objects with their textual positions, assisting GPT-4V to accurately identify and localize significant visual details.

\definecolor{darkgreen}{rgb}{0.0, 0.5, 0.0}
\begin{figure}[ht] % The placement specifier [ht] can be adjusted as needed
\centering
\begin{tcolorbox}
    \small 
    \includegraphics[width=\linewidth]{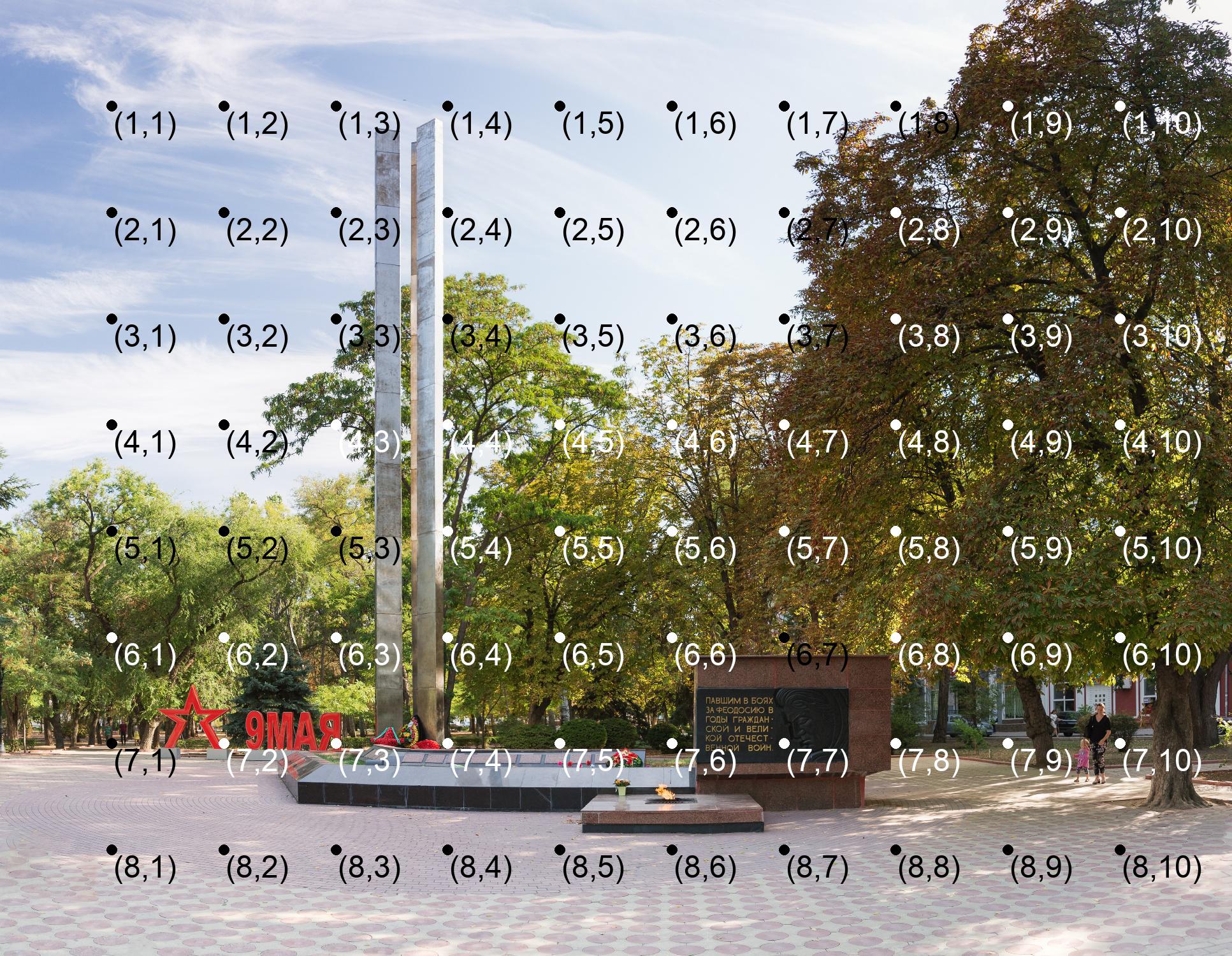} 
    \small
    \textbf{\textit{Question: What is the color of the little girl's shirt? Options: (A) pink. (B) white. (C) yellow. (D) black.}}

    \vspace{-1mm}
    \tcblower
    \small
    
    \textbf{Ground-Truth: } \textcolor{darkgreen}{A}

    \textbf{GPT-4V + CoT: } I'm sorry, but there seems to be a misunderstanding. \textcolor{red}{The image provided does not contain a little girl or any individuals wearing a shirt that can be described.} Therefore, I cannot provide an answer to the question about the color of a little girl's shirt based on this image.

    \textbf{GPT-4V + \method: } \textcolor{darkgreen}{The little girl is located near coordinates (7,9) and (7,10)}. Her shirt appears to be pink in color. Therefore, the answer is \textcolor{darkgreen}{A}.
\end{tcolorbox}
\vspace{-0.2cm}
\caption{An example from the V* Bench dataset.} 
\label{fig:vstar_example}
%\vspace{-0.5cm}
\end{figure}

\noindent\textbf{Fine-Grained Visual Understanding:} On benchmarks such as V* Bench~\cite{wu2023v} and Spotting Differences, \method has markedly improved the capabilities of LMMs in conducting visual search and localizing fine-grained visual details, with an average improvement of $11.7$. 
Taking Fig.~\ref{fig:vstar_example} as an example, 
with the support of coordinates, GPT-4V effectively pinpoints and documents the precise locations of target visual elements, contributing to its accurate perception of the target attribute. Additionally, we notice that without coordinates, GPT-4V is more easily to give up and apologize for its search failure.

\noindent\textbf{Hallucination:} With an average improvement of $5.7$, utilizing coordinates as a scaffold enables GPT-4V to recognize objects within a scene and further accurately describe their positions, guiding its textual reasoning to focus on precise visual information. With an example shown in Fig.~\ref{fig:pope_example} in Appendix~\ref{appendix:cases}, GPT-4V with coordinates is capable of precisely capturing visual details and preventing hallucinating non-existent objects, promoting accurate visual grounding.

\section{Ablation Study}
\label{section:ablations}

\begin{figure}[t]
    \centering
    \includegraphics[width=\linewidth]{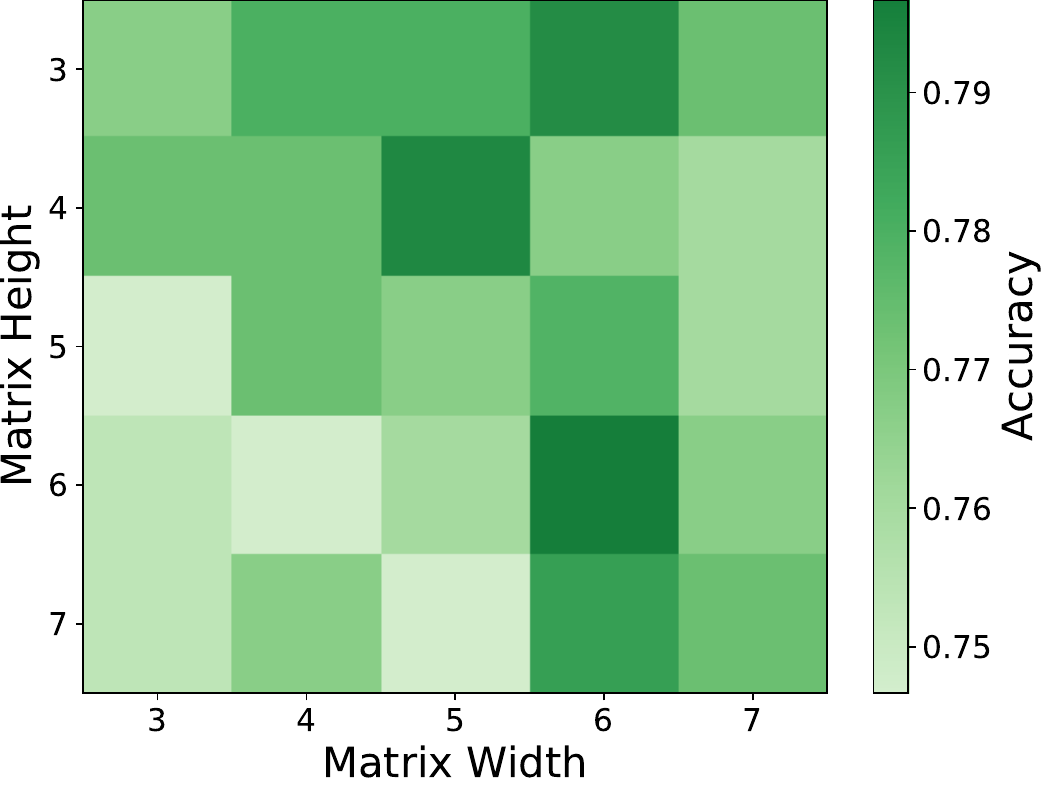}
    \vspace{-5mm}
    %\caption{Results of different matrix sizes in ablation experiments, with better accuracy illustrated in deeper green.}
    \caption{Impact of matrix sizes. Better accuracies are illustrated in darker green.}
    \label{fig:results_matrix_sizes}
    \vspace{-2mm}
\end{figure}

We conduct extensive ablation studies on key factors such as matrix size and coordinate color to validate and further explore \method.

\subsection{Experimental Setup}
Due to limited GPT-4V access quota, we each sample 50 questions from Visual Spatial Reasoning~\cite{liu2023vsr}, Winoground~\cite{thrush2022winoground}, and POPE (Adversarial Subset)~\cite{li2023evaluating}, creating an ablation subset of 150 samples. Overall accuracy per question is adopted as the metric and GPT-4V is used for our experiments. Additionally, for stable results, we run each experiment twice and report average accuracy.

%\subsection{Experiment 1. Matrix Size}
\subsection{Effect of Matrix Size}
\label{sec:matrix_size}
The matrix size $h$ and $w$ may influence the precision of textual reference and the granularity of visual information. Consequently, we incorporate matrices of difference sizes varying from $3\times3$ to $7\times7$ and measure their performances. 

Fig.~\ref{fig:results_matrix_sizes} depicts the performance variations with different matrix sizes, suggesting $6\times6$ as the optimal size for our ablation dataset.
Additionally, the sizes in the upper right section tend to perform better than those in the lower left section. It may be due to the sampled images usually possessing equal or larger widths than heights, suggesting matrix sizes may ideally align with image sizes.

Additionally, the $6\times6$ size did not yield the best results across all three subsets, hinting at potential improvements by customizing matrix sizes for specific tasks. It may be beneficial to automatically and dynamically adjust the matrix size and we leave this open problem to future research.

\begin{figure*}[h]
    \centering
    \begin{subfigure}{0.24\textwidth}
        \includegraphics[width=\textwidth]{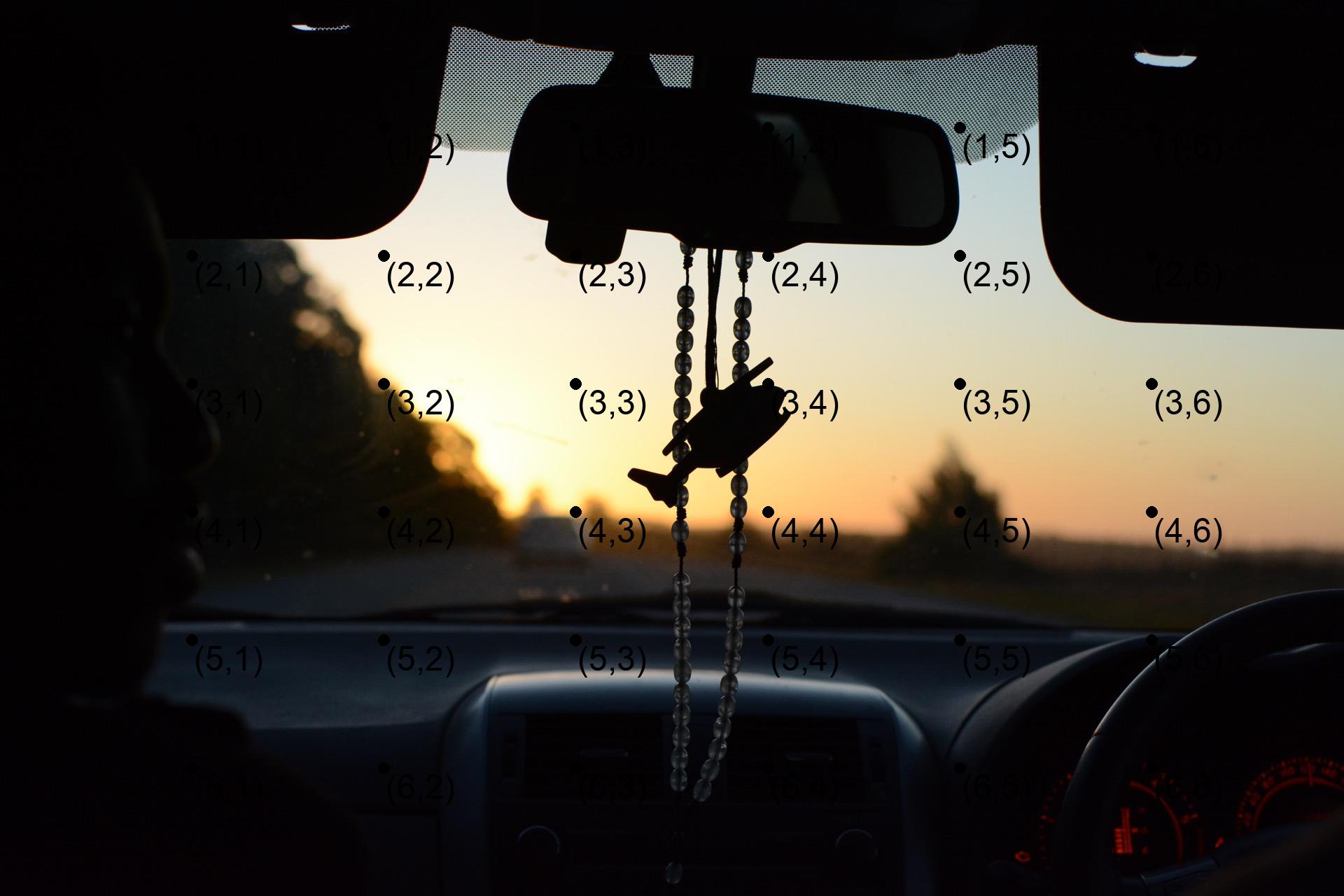}
        \caption{Black}
    \end{subfigure}\hfill
    \begin{subfigure}{0.24\textwidth}
        \includegraphics[width=\textwidth]{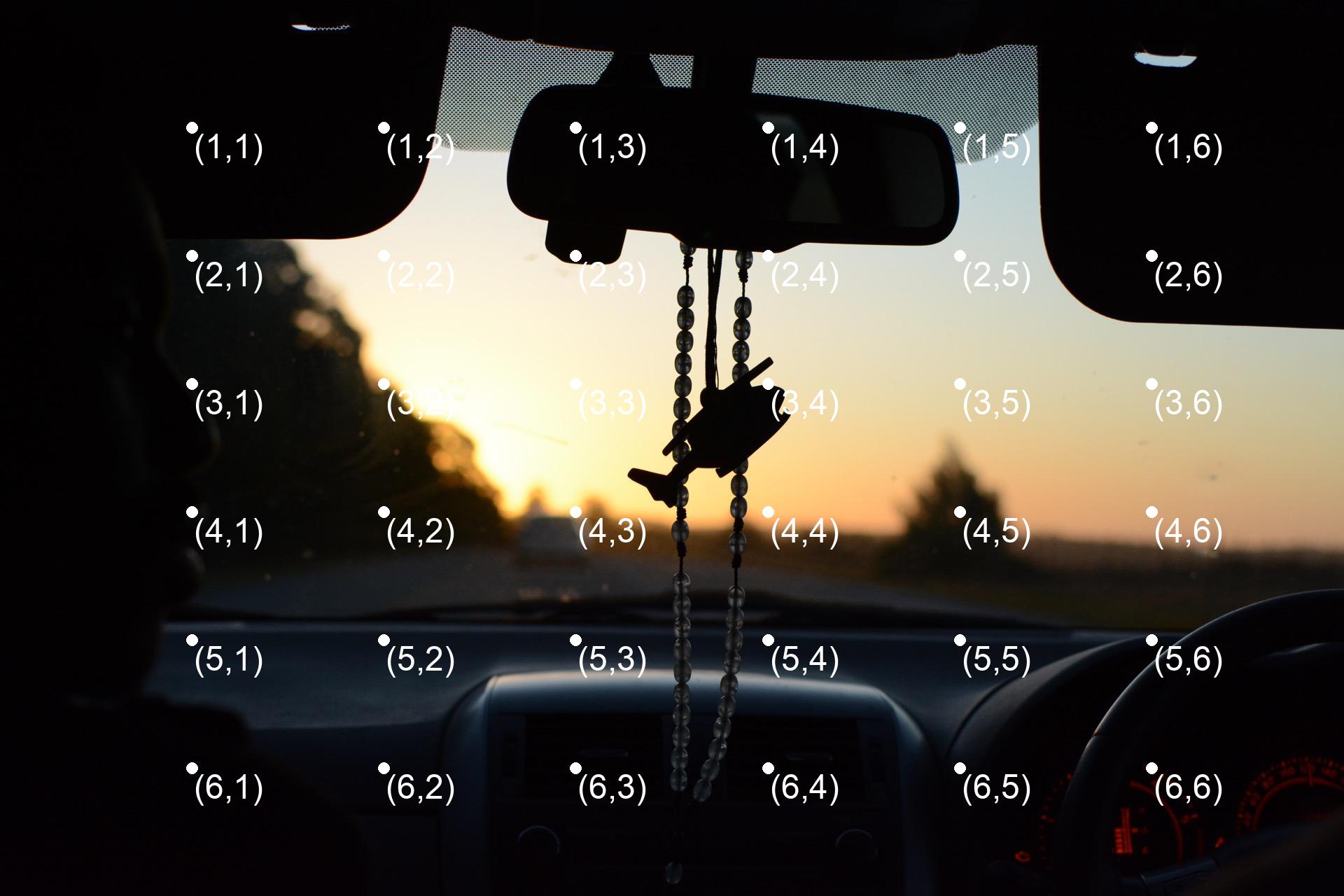}
        \caption{White}
    \end{subfigure}\hfill
    \begin{subfigure}{0.24\textwidth}
        \includegraphics[width=\textwidth]{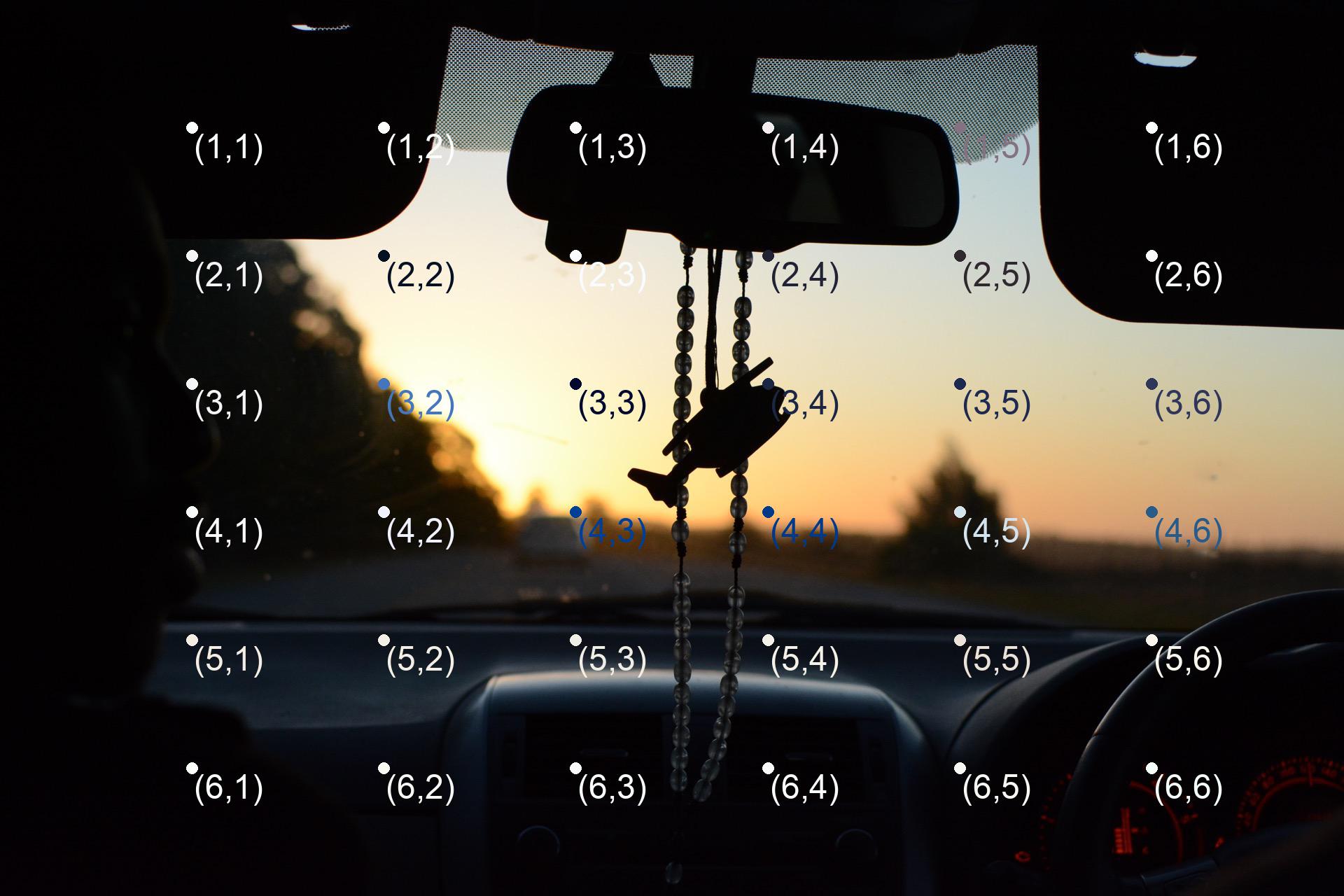}
        \caption{Complementary}
    \end{subfigure}\hfill
    \begin{subfigure}{0.24\textwidth}
        \includegraphics[width=\textwidth]{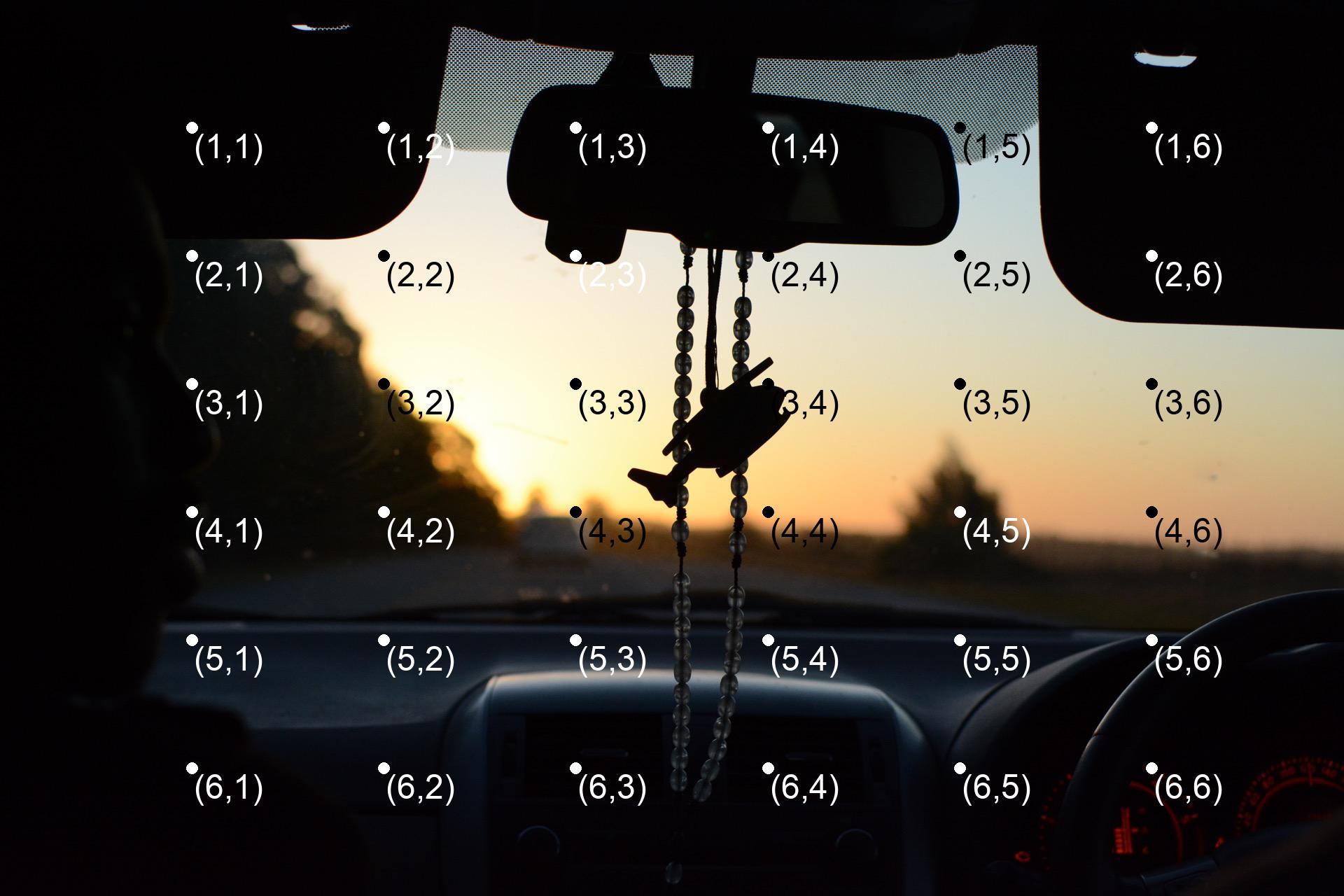}
        \caption{Binary (Ours)}
    \end{subfigure}\hfill
    \vspace{-2mm}
    \caption{Examples of different color configurations of dot matrices and coordinates.}
    \label{fig:color_comparision}
    \vspace{-4mm}
\end{figure*}

\begin{table}[]
\vspace{+0.2cm}
\resizebox{\linewidth}{!}{
\begin{tabular}{c|ccc|c}
\toprule
Coloring Strategy         & WNG    & VSR           & POPE    & Overall       \\ \midrule
None (Baseline)       & 70.0          & 64.0          & 73.0              & 69.0          \\
White                  & 78.0          & 72.0          & 78.0              & 76.0          \\
Black                  & 79.0          & \textbf{73.0} & 82.0              & 78.0          \\
Complementary          & 77.0          & 71.0          & 81.0              & 76.3          \\
Binary (Ours) & \textbf{81.0} & \textbf{73.0} & \textbf{84.0}     & \textbf{79.3} \\ \bottomrule
\end{tabular}
}
\caption{Results of different coloring strategies in ablation experiments,  where \textit{WNG} denotes the Winoground~\cite{thrush2022winoground} dataset and \textit{POPE} denotes the POPE~\cite{li2023evaluating} Adversarial subset.}
\label{tab:coloring}
% \vspace{-0.5cm}
\end{table}

\subsection{Effect of Matrix Color}
In terms of matrix color, we design different coloring strategies and compare their performances. As illustrated in Fig.~\ref{fig:color_comparision}, uniform coloring strategies adopt the same color for various scenes, occasionally blending into the surroundings. Complementary colors introduce large amounts of colors and may mislead model attention.
Consequently, for simplicity and visibility, we choose the most contrasting color from black and white at each dot location.
To assess the efficacy of our approach, we compared it against baseline coloring strategies, including \textit{uniform black}, \textit{uniform white}, and \textit{complementary coloring}. 
As shown in Table~\ref{tab:coloring}, our \textit{binary coloring} strategy slightly surpasses the alternatives in performance. 

\begin{table}[]
\vspace{+0.2cm}
\resizebox{\linewidth}{!}{
\begin{tabular}{c|ccc|c}
\toprule
Coordinates     & WNG    & VSR           & POPE & Overall       \\ \midrule
None (Baseline)        & 70.0          & 64.0          & 73.0              & 69.0          \\
Alphabet        & 80.0          & 72.0          & 79.0              & 77.0          \\
Pixel        & 78.0          & \textbf{75.0} & 77.0              & 76.7          \\
One-Dimensional & 72.0          & 71.0          & 81.0              & 74.7          \\
Cartesian (Ours)            & \textbf{81.0} & 73.0          & \textbf{84.0}     & \textbf{79.3} \\ \bottomrule
\end{tabular}
}
\caption{Results of different coordinates designs in ablation experiments, where \textit{WNG} denotes the Winoground~\cite{thrush2022winoground} dataset and \textit{POPE} denotes the POPE~\cite{li2023evaluating} Adversarial subset.}
%\vspace{-0.5cm}
\label{tab:coordinates}
\end{table}

\subsection{Effect of Coordinate Format}
Coordinates, as textual references for dots, are vital for aligning visual inputs with textual outputs. To assess the effectiveness of our implementation, we experiment with various coordinate formats, including alphabetic, one-dimensional numerical, and pixel absolute coordinates. The examples of these formats are exhibited in Fig.~\ref{fig:format_comparison} of Appendix~\ref{appendix:cases}.

The results, detailed in Table~\ref{tab:coordinates}, reveal that our approach surpasses alternative coordinate designs in performance. Additionally, all the coordinates designs perform better than the baseline without coordinates, indicating the flexibility of coordinates design and the adaptability to different scenarios.

\subsection{Effect of Dot Perturbations}
\begin{figure}[t]
    \centering
    \begin{subfigure}{0.48\linewidth}
        \includegraphics[width=\textwidth]{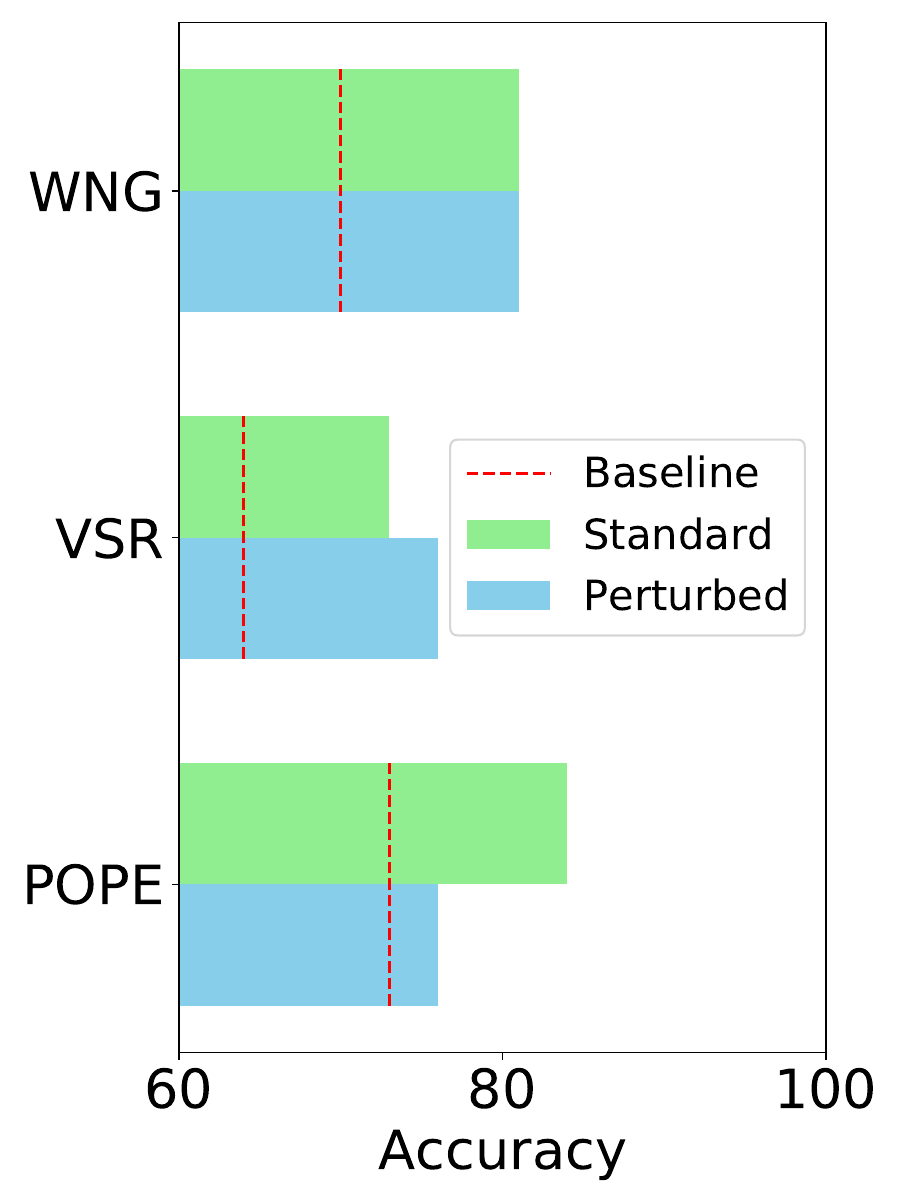}
    \end{subfigure}\hfill
    \begin{subfigure}{0.48\linewidth}
        \includegraphics[width=\textwidth]{figs/example_interrupt.jpg}
    \end{subfigure}
    \vspace{-2mm}
    \caption{The results of perturbed and standard coordinates (left) and a perturbed coordinates-overlaid example (right), where \textit{WNG} denotes the Winoground~\cite{thrush2022winoground} dataset and \textit{POPE} denotes the POPE~\cite{li2023evaluating} Adversarial subset.}\label{fig:interrupt_results}
    \vspace{-2mm}
\end{figure}

To assess the resilience of \method, we introduce Gaussian noise to the dots, slightly adjusting their positions without significantly changing their relative placements, as illustrated in Fig.~\ref{fig:interrupt_results}. We model the original dot positions as $(X, Y)$, with $l_h$ and $l_w$ representing the distances between neighboring dots along the $x$ and $y$ axes, respectively. The perturbed coordinate $(X_{\text{new}}, Y_{\text{new}})$ reads:

\vspace{-3mm}

%$$
\begin{equation}
\begin{bmatrix} \mathbf{X_{\text{new}}} \\ \mathbf{Y_{\text{new}}} \end{bmatrix} = \begin{bmatrix} \mathbf{X} \\ \mathbf{Y} \end{bmatrix} + \begin{bmatrix} \mathcal{N}\left(0, \left(\frac{1}{4} \cdot l_h\right)^2\right) \\ \mathcal{N}\left(0, \left(\frac{1}{4} \cdot l_w\right)^2\right) \end{bmatrix}
\end{equation}

According to the findings depicted in Fig.~\ref{fig:interrupt_results}, the coordinates perturbed with noise not only retain the enhancements provided by standard coordinates but also, in the VSR subset, exceed their performance. This indicates the substantial robustness against perturbation of the coordinates and suggests the potential for further optimizing their placement.

\section{Integration with Other Methods}
This section describes integration experiments of \method combined with active perception and Chain-of-Thought~\cite{wei2022chain}.

\begin{figure*}[t]
    \centering
    \includegraphics[width=\textwidth]{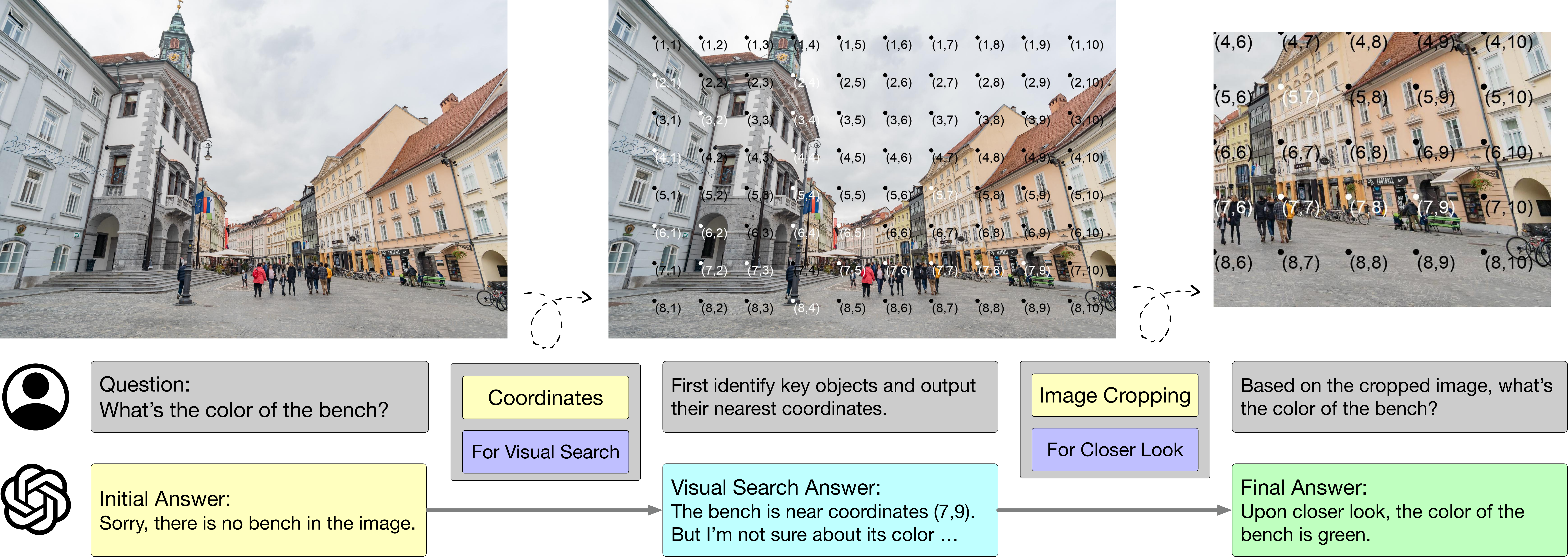}
    \caption{The procedure of combined \method and active perception techniques on V* Bench.}
    \label{fig:image_manipulation}
    \vspace{-0.2cm}
\end{figure*}

\subsection{\method + Active Perception}
In complex visual environments, humans would proactively engage with their surroundings to enhance scene understanding, like zooming in or changing perspectives. Similarly, we recognize that LMMs should possess such capabilities in realistic scenarios and propose that \method can function as a scaffold for effective active perception.

To validate this, we integrate \method with active perception in the experiments on the direct\_attributes subset of V* Bench~\cite{wu2023v}, which requires LMMs to perceive fine-grained details in high-resolution images. This challenge encompasses both the localization of target objects and the identification of their attributes under resolution constraints. Consequently, we adopt two metrics to measure LMM performance, including \textit{Not Found Rate} representing the percentage of invalid responses, and \textit{Success Rate} representing the percentage of correct responses. 

\begin{table}[]
\resizebox{\linewidth}{!}{
\begin{tabular}{c|cc}
\toprule
Method                   & N.F.R. $\downarrow$ (\%)  & S.R. $\uparrow$ (\%) \\ \midrule
Naive                    & 72.2                 & 21.7               \\
CoT                      & 71.3                 & 21.7               \\
\method                      & 26.2               & 31.3             \\
\method + A.P.   & \textbf{14.8}              & \textbf{45.2}             \\ \bottomrule
\end{tabular}
}
\vspace{-2mm}
\caption{Results of \method + active perception on V* Bench~\cite{wu2023v} direct\_attributes subset, where \textit{A.P.} denotes active perception, \textit{N.F.R.} denotes Not Found Rate and \textit{S.R.} denotes Success Rate.}
\label{tab:image_manipulation}
% \vspace{-5mm}
\end{table}

As depicted in Fig.~\ref{fig:image_manipulation}, our combined method unfolds in two phases: initial visual search to locate the target details, followed by cropping the image around the pinpointed coordinates to closely examine and identify the target attributes.

The results, presented in Table~\ref{tab:image_manipulation}, reveal a performance enhancement of 14.1\% compared with \method alone, underscoring the utility of coordinates in facilitating active perception. Furthermore, the results reveal two notable performance leaps. The initial improvement (CoT $\rightarrow$ \method) is attributed to the use of coordinates, significantly reducing the \textit{Not Found Rate} and thereby aiding in the visual search process. The subsequent gain (\method $\rightarrow$ \method + A.P.) results from the combined implementation of active perception, which enables LMMs to accurately discern target attributes within the cropped regions.

\begin{table}[]
\resizebox{\linewidth}{!}{
\begin{tabular}{c|cccc}
\toprule
\textbf{Dataset}                  & \textbf{Naive}               & \textbf{CoT}                 & \textbf{\method}                 & \textbf{\method + CoT} \\ \midrule
Wino.                           & 17.0 & 33.0 & 33.0 & \textbf{41.0}      \\
V*         & 27.2 & 30.8 & 44.6 & \textbf{47.9}      \\ \bottomrule
\end{tabular}
}
\vspace{-2mm}
\caption{Results of \method combined with Chain-of-Thought~\cite{wei2022chain} on V* Bench~\cite{wu2023v} and Winoground~\cite{thrush2022winoground}.}
\label{tab:cab_cot}
\vspace{-2mm}
\end{table}

\subsection{\method + Chain-of-Thought}

Our prompting method, characterized by its simplicity, can seamlessly integrate with zero-shot CoT by appending \textit{Let's think step by step.} to user instructions. To test its effectiveness, we conduct experiments on Winoground~\cite{thrush2022winoground} and V* Bench~\cite{wu2023v}. Results from Table~\ref{tab:cab_cot} demonstrate that combining our method with CoT enhances LMM performance beyond what either method achieves independently. These findings underscore our method's substantial compatibility and potential for performance improvement when combined with other methods.

\section{Conclusion}
In this work, we propose \method, a simple and general visual prompting method that utilizes scaffolding coordinates to promote vision-language coordination in LMMs. Extensive experiments show that \method successfully elicits LMM capabilities in several challenging vision-language tasks. 

\section*{Limitations}
Here we discuss two limitations of this work.

\noindent (1) To automatically adjust dot matrix attributes. In this work, for simplicity and clarity, we adopt matrices of size $6\times6$ in our implementation. However, our ablation study in Section~\ref{section:ablations} suggests that a one-size-fits-all matrix size can yield good, but not the best results across all datasets. Given the diversity of visual tasks and the varying granularity of information in different scenes, it stands to reason that tailoring the matrix attributes, such as size and coordinates format, to the specific requirements of each task or even each sample could improve performance. Addressing the dynamic and automatic adjustment of these attributes to better suit different scenarios remains an area for future exploration.

\noindent (2) To enhance precision in visual localization. By integrating dot matrices with coordinates onto images, we aimed to facilitate improved vision-language coordination by associating key objects with their closest coordinates. However, our observations indicate that, particularly in complex or clustered scenes, GPT-4V occasionally struggles to accurately associate textual reasoning with the nearest coordinates. This challenge underscores the need for LMMs to achieve improved visual localization and grounding capabilities in intricate environments. With \method, we expect the future of LMMs and visual prompting techniques to be further improved in terms of visual localization.

\bibliography{custom}
\bibliographystyle{acl_natbib}

\appendix

\section{Appendix}
\label{sec:appendix}
\subsection{Prompts}

This section exhibits the prompts used in \method implementation and our experiments.

\subsubsection{\method implementation}
\label{appendix:scaffold_prompts}
In \method implementation, we use the following textual guidelines to describe the effective use of coordinates.

\noindent\textbf{Single-Image Setting.} In the single-image setting, we label all the dots with two-dimensional coordinates and deliver both the original image and the coordinates-overlaid image to the model. Consequently, we use the following guidelines.

\begin{tcolorbox}
    \small
    I will provide you with two images of the same scene. The second image is overlaid with a dot matrix of the shape of HEIGHT * WIDTH to help you with your task, and each dot is labeled with two-dimensional coordinates (x,y).

    1. When you mention any key objects in the image, first output their nearest coordinates then identify them.
    
    2. You use the coordinates to determine the spatial relationships of the objects. Within each column, the x-coordinate increases from top to bottom, and within each row, the y-coordinate increases from left to right.
    
    3. You can search and reason region by region with the help of the dots.
    
    4. Finally, conclude your answer in format [[ANSWER]], such as [[A]], [[B]], [[C]] or [[D]].
    
\end{tcolorbox}

Note that the fourth guideline serves as a constraint for specific output formats and may vary among different tasks.

\noindent\textbf{Double-Images Setting.} In a double-images setting, we label all the dots with three-dimensional coordinates, with the first coordinate serving to distinguish between two images. Consequently, we use the following guidelines.

\begin{tcolorbox}
    \small
    Two images are provided, each overlaid with a grid of dots arranged in a matrix with dimensions {h} by {w}. Each dot on this grid is assigned a unique set of three-dimensional coordinates labeled as (t, x, y). The first coordinate, 't', serves to distinguish between the two images: '1' is assigned to the first image on the left, and '2' to the second image on the right. The other two coordinates, 'x' and 'y', are used to specify the dot's spatial location within its respective image. This labeling system is designed to assist you in identifying and referring to specific points within each image.
    
    1. When you mention any key objects in the image, first output their nearest coordinates then identify them.
    
    2. You use the coordinates to determine the spatial relationships of the objects. within each column, the x-coordinate increases from top to bottom, and Within each row, the y-coordinate increases from left to right.
    
    3. You can search and reason region by region with the help of the dots.
    
    4. Finally, you must conclude your answer in format [[ANSWER]], such as [[A]] or [[B]].
    
\end{tcolorbox}

Note that the fourth guideline serves as a constraint for specific output formats and may vary among different tasks.

\noindent\textbf{Image-Sequence Setting.} In the image sequence setting, we label all the dots with three-dimensional coordinates. For simplicity and efficiency, we only deliver the coordinates-overlaid image to the model. Consequently, we use the following guidelines.

\begin{tcolorbox}
    \small
    A sequence of images is provided, each overlaid with a grid of dots arranged in a matrix with dimensions HEIGHT by WIDTH. Each dot on this grid is assigned a unique set of three-dimensional coordinates labeled as (t, x, y). The first coordinate, 't', serves to distinguish between the images, for instance, '1' is assigned to the first image, and '8' to the last image. The other two coordinates, 'x' and 'y', are used to specify the dot's spatial location within its respective image. This labeling system is designed to assist you in identifying and referring to specific points within each image.
    
    1. When you mention any key objects in the image, first output their nearest coordinates then identify them.
    
    2. You use the coordinates to determine the temporal and spatial relationships of the objects. Within the image sequence, the t-coordinate increases as time grows; within each column, the x-coordinate increases from top to bottom; within each row, the y-coordinate increases from left to right.
    
    3. You can search and reason region by region with the help of the dots.
    
    4. you need to keep your descriptions concise and clear.
    
\end{tcolorbox}

Note that the fourth guideline serves as a constraint for specific output formats and may vary among different tasks.

\subsubsection{Datasets}
\label{appendix:datasets}

Spotting Differences dataset is our newly formulated dataset challenging LMMs to precisely localize different details in two similar images. We exhibit the prompts as follows.

\noindent\textbf{Spotting Differences~\footnote{https://www.crazygames.com/game/find-the-difference}}
\begin{tcolorbox}
    \small 
    \includegraphics[width=\linewidth]{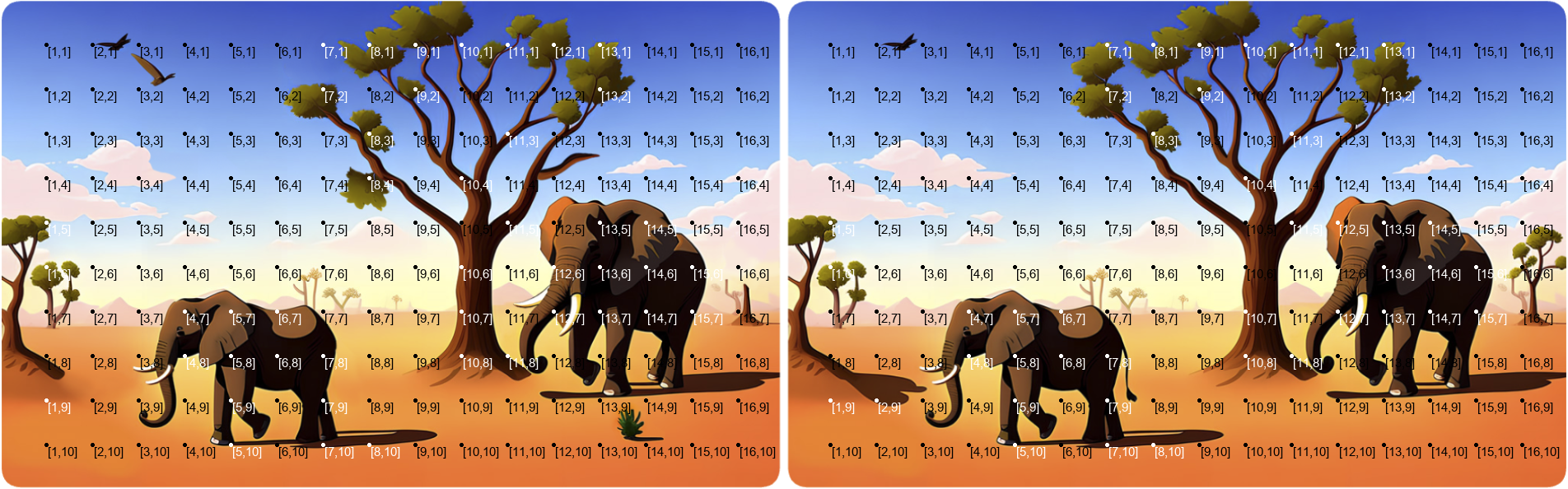} 

    \tcblower
    \small
[Question] 
Spot ten differences between the images and answer the dot position closest to every difference in the format: [spot index, x, y]. I am going to tip \$100 for a better answer.

[Answer]
Let's identify ten differences between the left and right images:

1. The bird in the sky is missing in the right image. Closest dot: [1, 2, 1]...
    
\end{tcolorbox}

\subsubsection{Integration Experiments}

We list the prompts used in the integration experiments.

\noindent\textbf{\method + Active Perception}
Firstly, we guide the LMM to visually search the image and localize target objects using the following prompt.

\begin{tcolorbox}
    \small
    Based on the question: {question} you should first identify key objects in the question and link them with their nearest coordinate, and finally conclude the coordinates in format [[(x,y)]] in the end(you don't need to answer the question).
    
\end{tcolorbox}

Secondly, we crop the image based on the output coordinates and guide the LMM to answer the question based on the cropped images in the second turn of the conversation.

\begin{tcolorbox}
    \small
    Here are the cropped images from the scene according to your selected coordinates, you can take a closer look and answer the question. If I don't provide cropped images or the target does not exist in the cropped image, please visually search the original image and answer the question. 
    
    Question: {QUESTION}  
    
    Options: {OPTIONS}
    
\end{tcolorbox}

\subsection{Benchmarks}
This section details the benchmarks used in our experiments to evaluate our method. The benchmarks are divided into four categories and elaborated respectively.

\subsubsection{Spatial Reasoning}
Spatial reasoning is a crucial capability of LMMs to determine spatial relationships between objects in the image. To evaluate the effectiveness of our method to elicit spatial reasoning capabilities, we select several challenging datasets including MME~\cite{fu2023mme} Position split, Visual Spatial Reasoning~\cite{liu2023vsr} dataset, EgoThink~\cite{cheng2023visionlanguage} Spatial split. Note that we only select a subset related to spatial reasoning in comprehensive evaluation benchmarks like MME~\cite{fu2023mme} and EgoThink~\cite{cheng2023visionlanguage} because the other subsets don't constitute spatial reasoning and our GPT-4V access quota is limited. The details of the benchmarks are as follows.

\noindent\textbf{MME (Position)~\cite{fu2023mme}}

\noindent\textit{\textbf{Dataset Introduction.}} The MME~\cite{fu2023mme} benchmark is a comprehensive evaluation suite designed to measure the perception and recognition capabilities of LMMs in 14 tasks. To evaluate spatial reasoning capabilities, we only select Position split from 14 subtasks. It contains 60 questions about the spatial relationship among the objects in the scene, such as \textit{left}, \textit{above}, etc.. 

\noindent\textit{\textbf{Metric.}} All questions are formatted as general interrogative sentences, which can be answered with either "yes" or "no". Therefore, we guide the model to output the final answer within double square brackets [[]], and then use accuracy as the metric to evaluate the model's performance.

\noindent\textit{\textbf{Example.}}

\begin{tcolorbox}
    \small 
    \includegraphics[width=0.8\linewidth]{figs/appendix/mme_example.jpg} 

    \tcblower
    \small

    Is the cricket bat above the batter's body? Please answer yes or no.
    
\end{tcolorbox}

\noindent\textbf{Visual Spatial Reasoning (VSR)~\cite{liu2023vsr}}

\noindent\textbf{\textit{Dataset Introduction.}} The VSR~\cite{liu2023vsr} dataset is designed to comprehensively evaluate LMM capabilities to perform spatial reasoning in images, containing 66 types of spatial descriptions in language. 
Each sample provides an image and a corresponding spatial description in terms of two individual objects in the scene. Because of the limited GPT-4V~\cite{openai2023gpt4} access quota, we randomly sample 200 samples from the dataset for evaluation.

\noindent\textit{\textbf{Metric.}} The task is to determine the correctness of the given spatial description, which can be answered with either "true" or "false". Therefore, we guide the model to output the final answer within double square brackets [[]], and then use accuracy as the metric to evaluate the model's performance.

\noindent\textit{\textbf{Example.}}

\begin{tcolorbox}
    \small 
    \includegraphics[width=0.8\linewidth]{figs/appendix/vsr_example.jpg} 

    \tcblower
    \small

    Determine whether the following statement is true or false: The person is facing the banana.
    
\end{tcolorbox}

\noindent\textbf{EgoThink (Spatial)~\cite{cheng2023visionlanguage}}

\noindent\textbf{\textit{Dataset Introduction.}} The Egothink~\cite{cheng2023visionlanguage} dataset is intended for first-perspective vision-language problem-solving capabilities. To evaluate spatial reasoning capabilities, we only select Spatial split from it. The Spatial split contains 50 questions about the spatial relationship among the objects, particularly requiring LMMs to perceive and reason from the first perspective. 

\noindent\textit{\textbf{Metric.}} The task is to answer spatial questions from the first perspective. We adopt the evaluation script in EgoThink official implementation~\footnote{https://github.com/AdaCheng/EgoThink}, which uses GPT-4~\cite{openai2023gpt4} as the judge model to score the answers.

\noindent\textit{\textbf{Example.}}

\begin{tcolorbox}
    \small 
    \includegraphics[width=\linewidth]{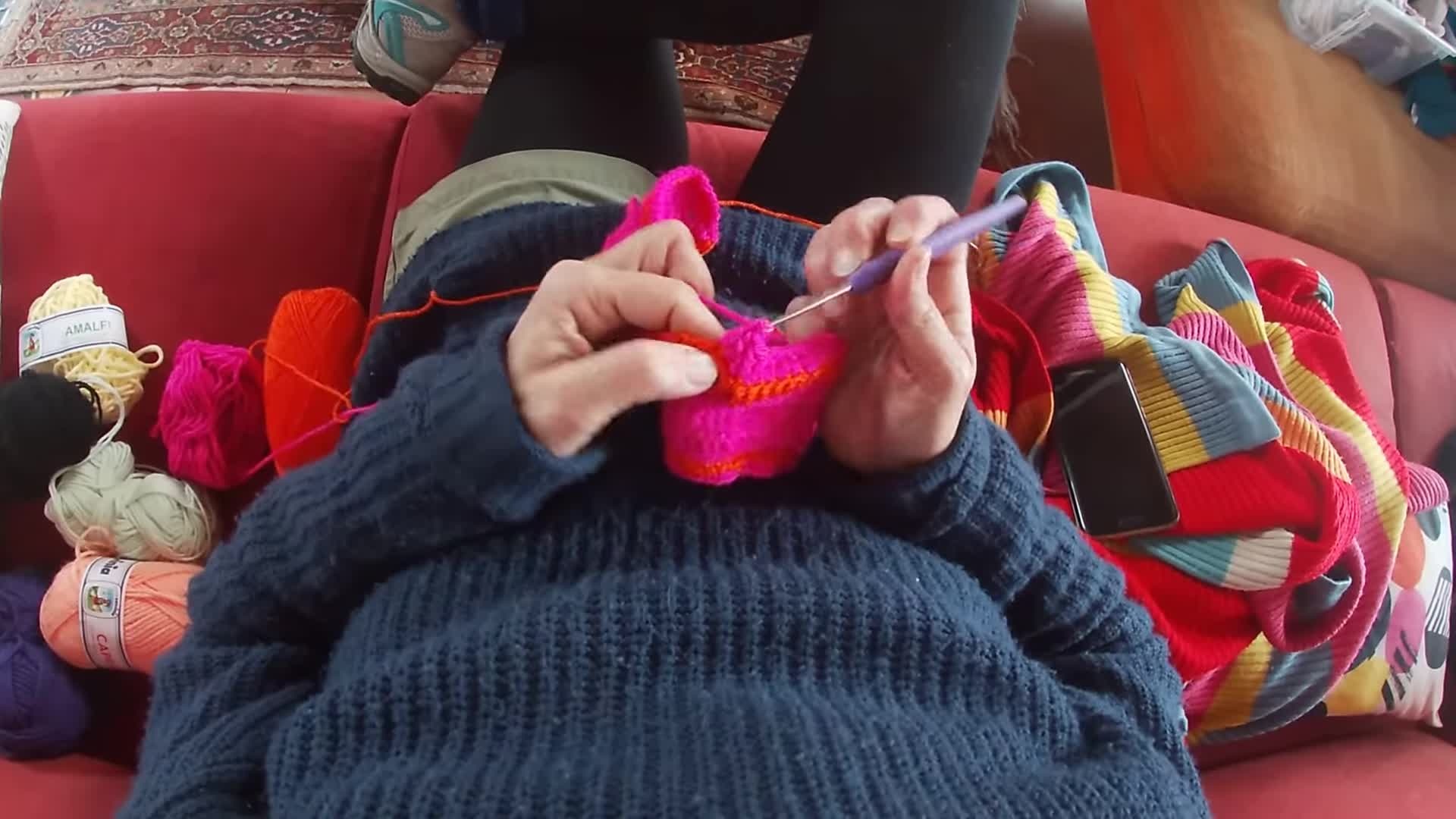} 

    \tcblower
    \small

    Where is the phone, on my left or on my right?
    
\end{tcolorbox}

\subsubsection{Compositional Reasoning}
Compositional reasoning represents the capability of LMMs to perceive and reason in terms of objects' attributes and their relationships, significant for visual perception. To evaluate the effectiveness of our method to elicit compositional reasoning capabilities, we select several challenging datasets including Winoground~\cite{thrush2022winoground}, WHOOPS!~\cite{bittonguetta2023breaking} and CLEVR~\cite{johnson2016clevr}. 

\noindent\textbf{Winoground~\cite{thrush2022winoground}}

\noindent\textbf{\textit{Dataset Introduction.}} Winoground~\cite{thrush2022winoground} proposes a novel dataset that challenges LMMs to correctly match two images and two captions. The captions use the same words but in a different order, requiring a precise understanding of both images and captions. The challenging dataset is a suitable benchmark for compositional reasoning.

\noindent\textit{\textbf{Metric.}} Given two images and two captions, we compose a sample into four binary-choice questions for the effective evaluation of LMMs, including choosing the correct caption given two images and choosing the correct image given two captions respectively. Finally, we adopt \textit{group score} to measure LMM performance: only when the model answers all four questions correctly is it considered completely correct for this sample.

\noindent\textit{\textbf{Example.}}

\begin{tcolorbox}
    \includegraphics[width=0.45\linewidth]{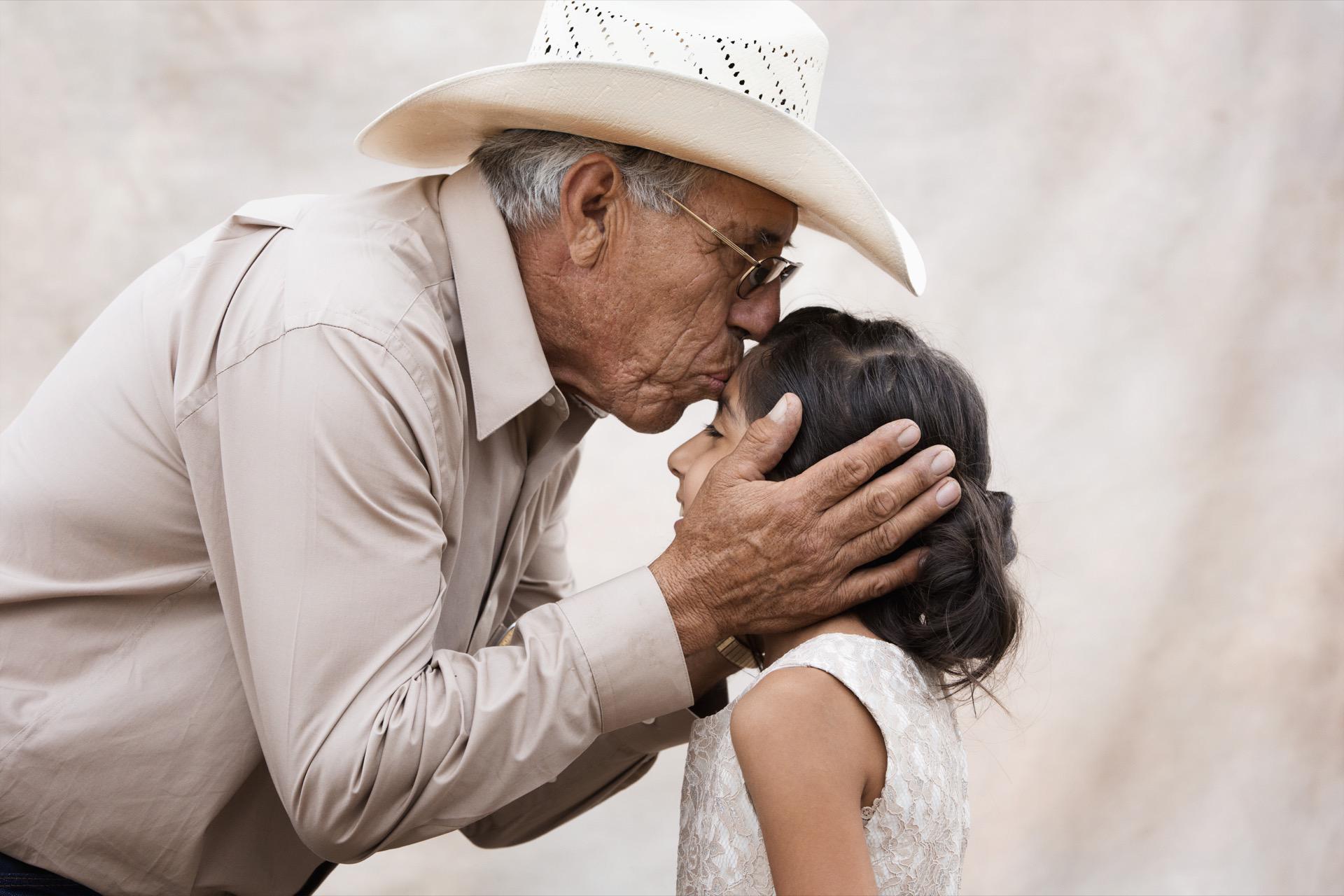}\hfill
    \includegraphics[width=0.45\linewidth]{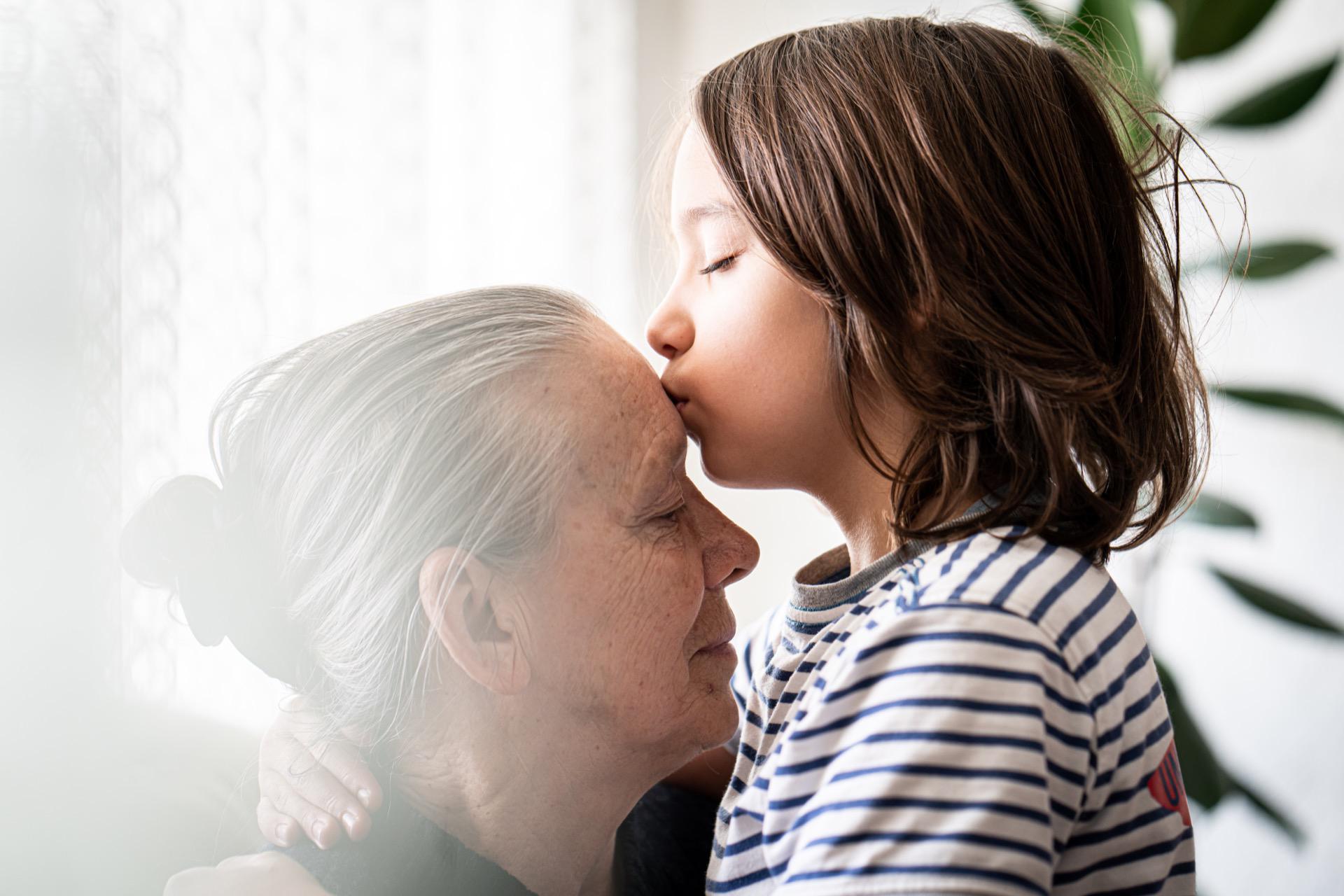} 

    \tcblower
    \small

    Q1: \textcolor{blue}{[[Image1]] [[Image2]] } Choose the correct image for the caption. Caption: an old person kisses a young person. Options: (A) image 1(left) (B) image 2(right)

    Q2: \textcolor{blue}{[[Image1]] [[Image2]] } Choose the correct image for the caption. Caption: a young person kisses an old person. Options: (A) image 1(left) (B) image 2(right) 

    Q3: \textcolor{blue}{[[Image1]]} Choose the correct caption for the image. (A) an old person kisses a young person. (B) a young person kisses an old person.

    Q4: \textcolor{blue}{[[Image2]]} Choose the correct caption for the image. (A) an old person kisses a young person. (B) a young person kisses an old person. 
    
\end{tcolorbox}

\noindent\textbf{WHOOPS!~\cite{bittonguetta2023breaking}}

\noindent\textbf{\textit{Dataset Introduction.}} The WHOOPS!~\cite{bittonguetta2023breaking} is designed to challenge LMMs to perform compositional reasoning in terms of purposefully commonsense-defying images. It remains challenging for LMMs to recognize and interpret these unconventional images. Furthermore, several tasks are posed over the dataset, and we select the visual question-answering task.

\noindent\textit{\textbf{Metric.}} We adopt the BEM score in WHOOPS! official implementation~\footnote{https://whoops-benchmark.github.io/} to evaluate LMM performances.

\noindent\textit{\textbf{Example.}}

\begin{tcolorbox}
    \small 
    \includegraphics[width=0.8\linewidth]{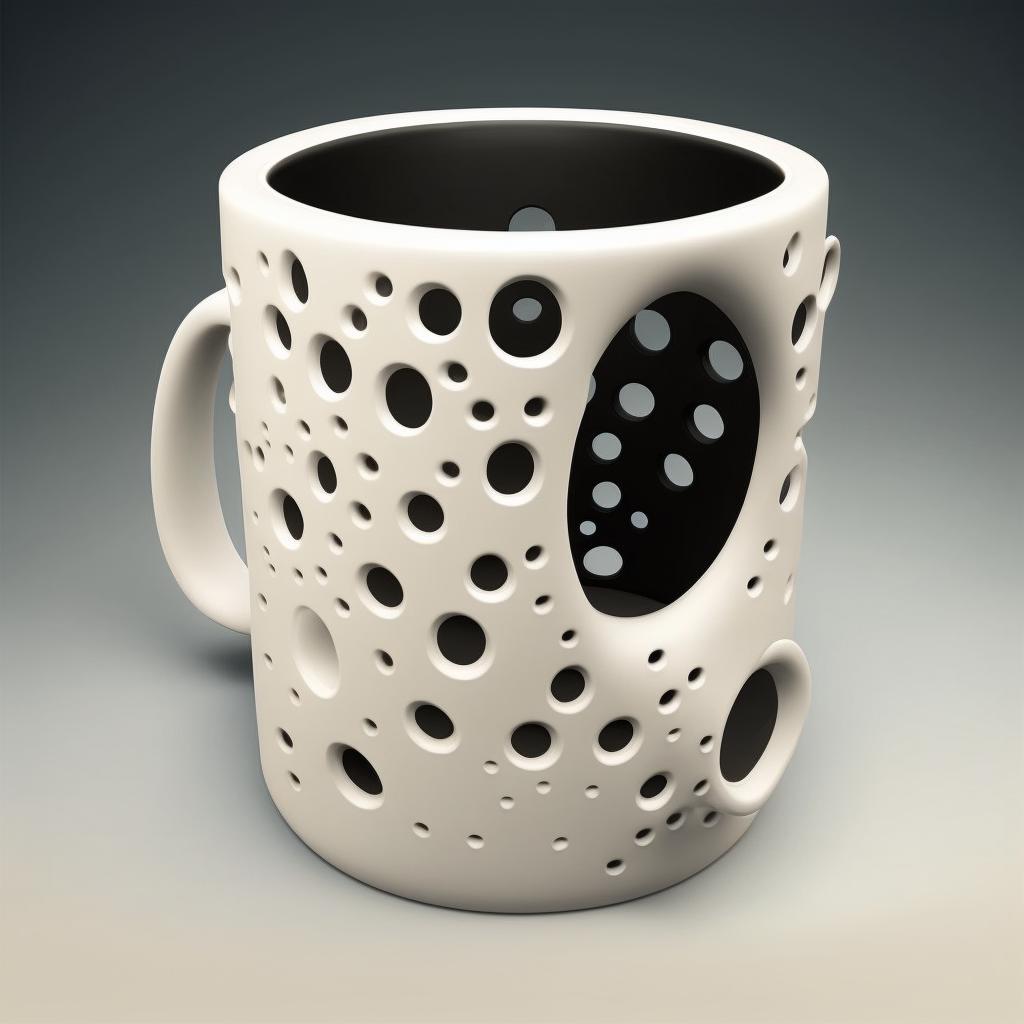} 

    \tcblower
    \small

    What is on a table with holes through the material?
    
\end{tcolorbox}

\noindent\textbf{CLEVR~\cite{johnson2016clevr}}

\noindent\textbf{\textit{Dataset Introduction.}} The CLEVR~\cite{johnson2016clevr} is designed as a standard evaluation suite for compositional vision-language reasoning with program-rendered images and correctly generated annotations. Containing objects like a metal cube or red sphere, the dataset poses questions in terms of object existence, object attributes, and object relationships. For effective evaluation, we randomly sampled 200 samples in the dataset.

\noindent\textit{\textbf{Metric.}} We guide the LMMs to answer the questions in an open-ended generative manner. Due to the complexity of the question and the answer, we adopt GPT-4~\cite{openai2023gpt4} as a judge model to determine the correctness of answers. Our judge prompt is as follows, adapted from MT-Bench~\cite{zheng2023judging}.

\begin{tcolorbox}
    \small

    [Instruction] Please act as an impartial judge and evaluate the quality of the response provided by an AI assistant to the user question displayed below. Your evaluation should consider correctness and helpfulness. You will be given a reference answer and the assistant's answer. Begin your evaluation by comparing the assistant's answer with the reference answer. Identify and correct any mistakes. The assistant has access to an image along with questions but you will not be given images. Therefore, please consider only how the answer is close to the reference answer. If the assistant's answer is not exactly the same as or similar to the answer, then he must be wrong.  Be as objective as possible. Discourage uninformative answers. Also, equally, treat short and long answers and focus on the correctness of answers.  After providing your explanation, you must rate the response with either 0, 0.5, or 1 by strictly following this format: "[[rating]]", for example: "Rating: [[0.5]]".
    
    [Question] \textcolor{blue}{question} 
    
    [The Start of Reference Answer] \textcolor{blue}{ground\_truth} [The End of Reference Answer] 
    
    [The Start of Assistant's Answer] \textcolor{blue}{answer} [The End of Assistant's Answer]
    
\end{tcolorbox}

\noindent\textit{\textbf{Example.}}

\begin{tcolorbox}
    \small 
    \includegraphics[width=\linewidth]{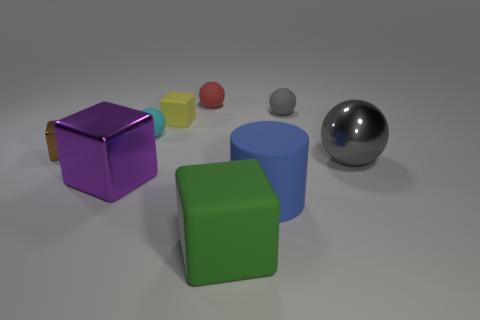} 

    \tcblower
    \small

    Are there more large balls behind the tiny brown cube than green shiny cylinders?
    
\end{tcolorbox}

\subsubsection{Fine-Grained Visual Understanding} Fine-grained Visual Understanding represents the capability of LMMs to precisely capture, perceive, and describe certain visual details in the scene. To evaluate the effectiveness of our method to elicit Fine-grained Visual Understanding, we select several challenging datasets including V* Bench~\cite{wu2023v} and Spotting Differences. 

\noindent\textbf{V* Bench~\cite{wu2023v}}

\noindent\textbf{\textit{Dataset Introduction.}} The V* Bench~\cite{wu2023v} is designed to challenge LMMs to perform visual search and identify fine-grained visual details in high-resolution images. The challenges lie in the necessity of visual search, while GPT-4V sometimes fails to capture visual details, responding with \textit{I'm sorry, I couldn't find XXX in the image}.
Note that we conduct our evaluation in an open-ended generative manner, GPT-4V sometimes fails to identify targets and refuses to choose an option.

\noindent\textit{\textbf{Metric.}} The questions are multiple-choice questions. Due to the generative nature of current LMMs, we pose a question to the target LMM and let it generate an open-ended response. Finally, we guide the model to output the final answer within double square brackets [[]], and then use accuracy as the metric to evaluate the model's performance.

\noindent\textit{\textbf{Example.}}

\begin{tcolorbox}
    \small 
    \includegraphics[width=\linewidth]{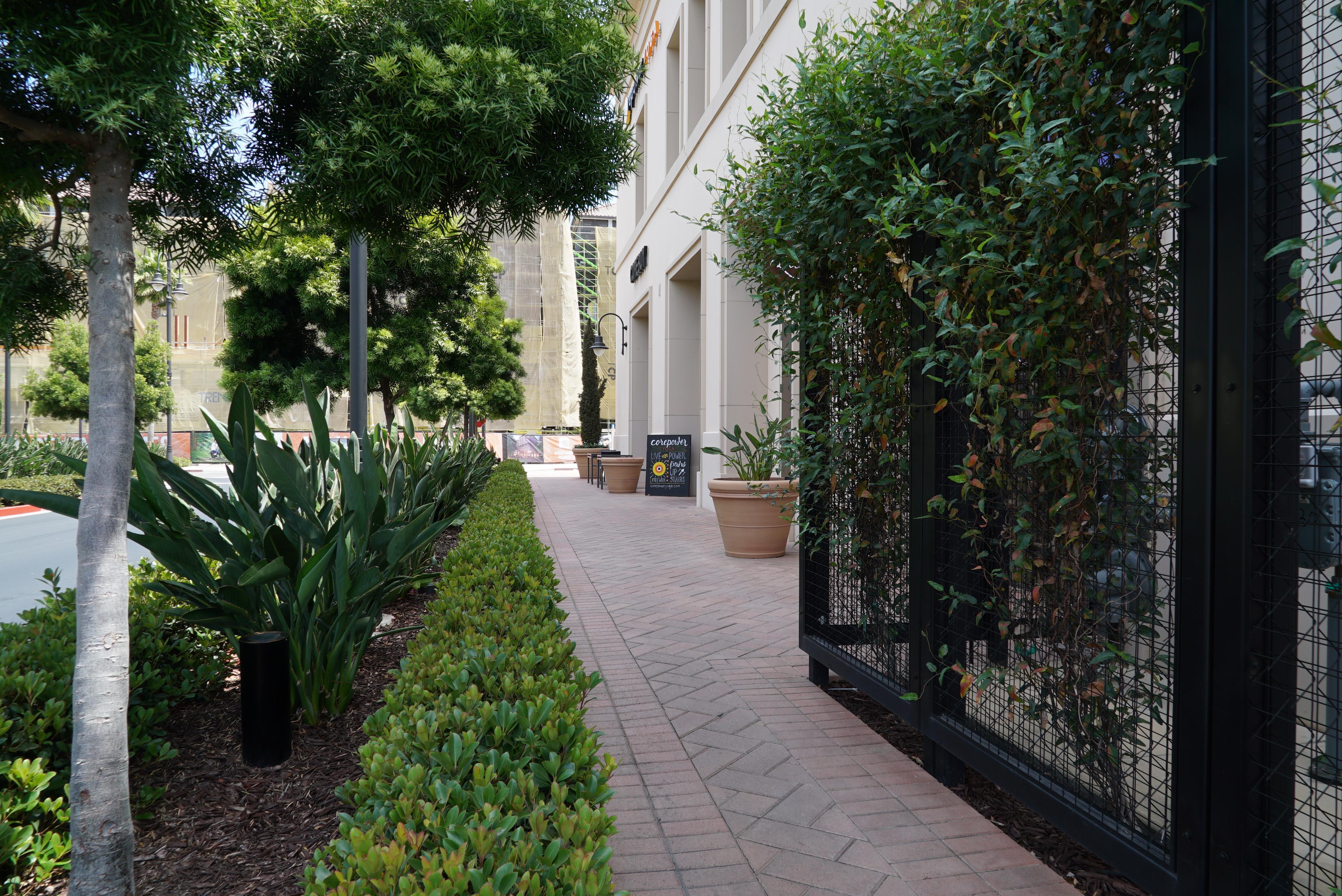} 

    \tcblower
    \small

    From the information on that advertising board, what is the type of this shop? 
    
    Options: (A) The shop is a yoga studio. (B) The shop is a cafe. (C) The shop is a seven-eleven. (D) The shop is a milk tea shop.
    
\end{tcolorbox}

\noindent\textbf{Spotting Differences~\footnote{https://www.crazygames.com/game/find-the-difference}}
\label{appendix:spotting_differences}

\noindent\textbf{\textit{Dataset Introduction.}} The dataset was designed to challenge the spatial analysis capabilities and object localization abilities of LLMs. Inspired and derived by the "Spot the Difference" web game, each level consists of two images. The goal is to locate the differences between the two images at 10 specific locations. The first 50 levels, increasing in difficulty, were selected as samples for this dataset. The challenge lies in the LLM's ability to first analyze the images and then accurately localize the correct object positions. GPT-4V may sometimes return \textit{I'm sorry, but I cannot assist with this request.}. 

\noindent\textit{\textbf{Metric.}} We guide GPT-4V to answer the pixel or matrix positions of the difference by dividing the prompt into question text + standardized answer requirement + prompt trick. We use OpenCV's Hough Circle Transform to locate the correct image position of the difference. When the distance between the correct position and the position given by GPT-4V is less than 50 pixels, the difference positioning is considered successful. 

\noindent\textit{\textbf{Example.}}

\begin{tcolorbox}
    \small 
    \includegraphics[width=\linewidth]{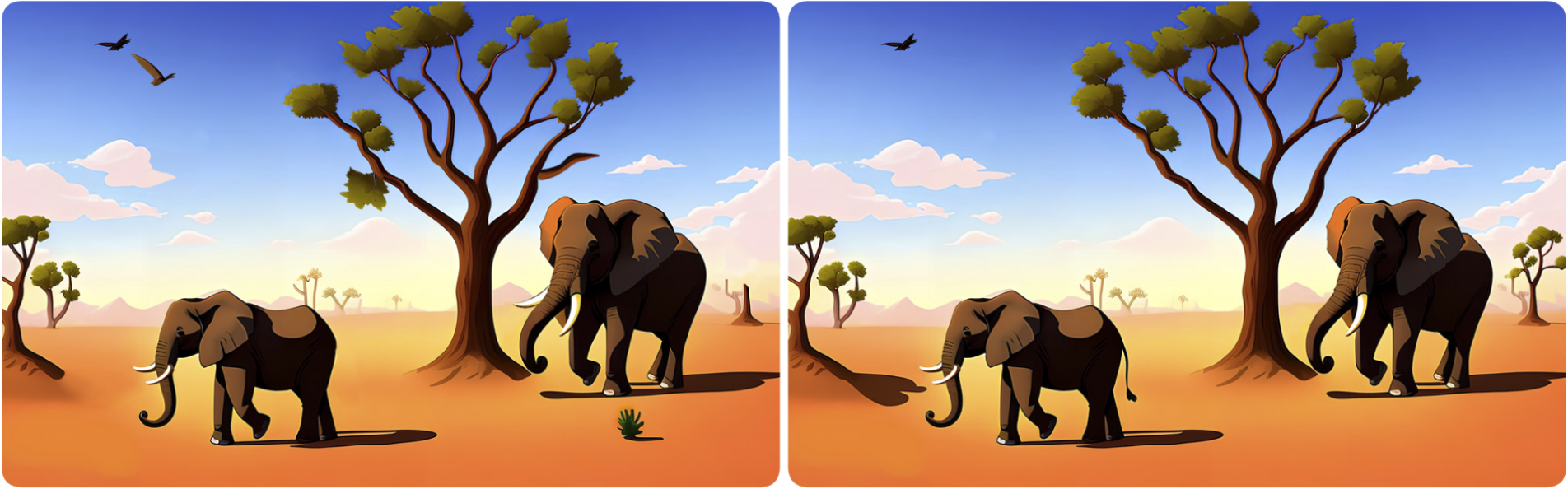} 
    \textbf Question Image.
    
    \includegraphics[width=\linewidth]{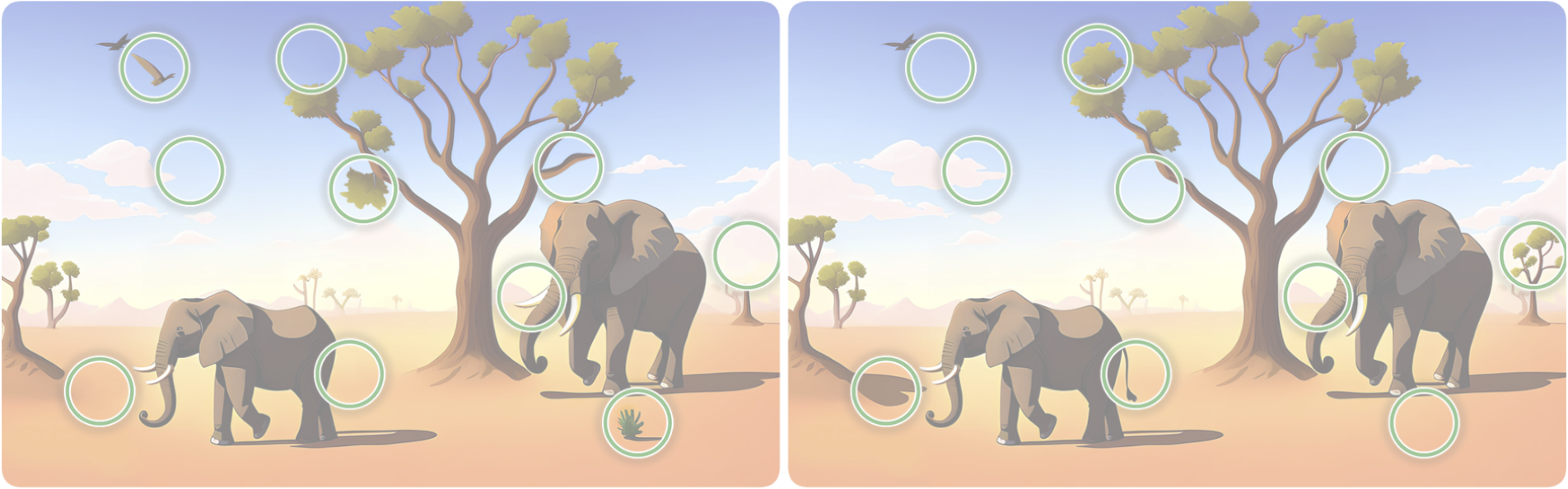} 
    \textbf Correct answer Image.
    \tcblower
    \small

[Question] 
Spot ten differences between the images and give mathematic coordinates of every difference to make sure I can locate the image with my program. The coordinate can be that the left edge of the left image is x=0, the right edge of the right image is x=1904 (based on the resolution of the image), and the top and bottom edges of the images are y=0 and y=594. please note that you only need to indicate the coordinate in the left image for difference because I want to use coordinates in only one image use [spot index, x, y], do not indicate coordinate in the right image(it means that the x of coordinate is less than 1000). I am going to tip \$100 for a good solution. 

[Answer]
Here are the ten differences between the two images along with their approximate coordinates on the left image:

1. [1, 150, 100] - The bird flying in the sky is missing...
    
\end{tcolorbox}

\subsubsection{Hallucination}
Hallucination in vision-language contexts represents a misalignment between visual inputs and textual outputs in terms of object existence, attributes, and relationships. To evaluate the effectiveness of our method to mitigate hallucination in vision-language tasks, we select several challenging datasets including POPE Adversarial subset~\cite{li2023evaluating}, HallusionBench (Hard)~\cite{guan2023hallusionbench} and Mementos~\cite{wang2024mementos}. The details of these benchmarks are as follows.

\noindent\textbf{POPE~\cite{li2023evaluating}}

\noindent\textit{\textbf{Dataset Introduction.}} The POPE~\cite{li2023evaluating} benchmark is designed for measuring object hallucination in images. The questions are built based on the object annotations, challenging LMMs to determine object existence in images.

\noindent\textit{\textbf{Metric.}} All questions are formatted as general interrogative sentences, which can be answered with either "yes" or "no". Therefore, we guide the model to output the final answer within double square brackets [[]], and then use accuracy as the metric to evaluate the model's performance.

\noindent\textit{\textbf{Example.}}

\begin{tcolorbox}
    \small 
    \includegraphics[width=\linewidth]{figs/appendix/pope_example.jpg} 

    \tcblower
    \small

    Is there a tennis racket in the image?
    
\end{tcolorbox}

\noindent\textbf{HallusionBench~\cite{li2023evaluating}}

\noindent\textit{\textbf{Dataset Introduction.}} The HallusionBench~\cite{li2023evaluating} is designed for evaluating language hallucination and visual illusion in LMMs. The images and questions are meticulously crafted by human experts, posing great challenges to current LMMs. Due to the limited GPT-4V quota, we evaluate our method on the \textit{Hard} set.

\noindent\textit{\textbf{Metric.}} All questions are formatted as general interrogative sentences, which can be answered with either "yes" or "no". Therefore, we simplify the evaluation process~\footnote{the official evaluation process of HallusionBench involves GPT-4 as judge model} and guide the model to output the final answer within double square brackets [[]], and then use accuracy as the metric to evaluate the model's performance.

\noindent\textit{\textbf{Example.}}

\begin{tcolorbox}
    \small 
    \includegraphics[width=\linewidth]{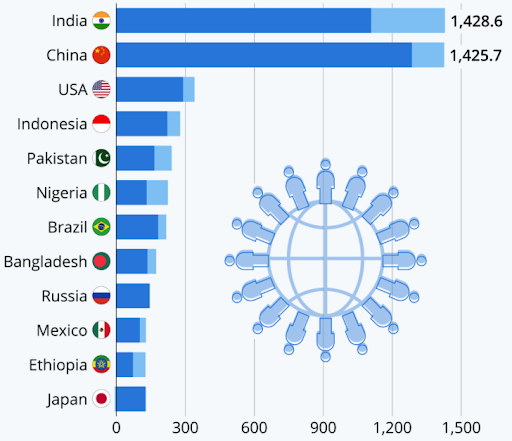} 

    \tcblower
    \small

    According to the chart, does China have the second largest population in the world?
    
\end{tcolorbox}

\noindent\textbf{Mementos~\cite{wang2024mementos}}

\noindent\textit{\textbf{Dataset Introduction.}} The Mementos~\cite{wang2024mementos} is designed for evaluating reasoning capabilities across image sequences. Featuring 4,761 diverse image sequences with varying lengths, the dataset adopts a GPT-4\_assisted metric to evaluate the correctness of objects and behaviors in generated descriptions, reflecting both reasoning capabilities and hallucination levels.

\noindent\textit{\textbf{Metric.}} The task is to generate a description of the image sequences. We adopt the evaluation script in Mementos official implementation~\footnote{https://github.com/umd-huang-lab/Mementos}, which uses GPT-4~\cite{openai2023gpt4} as the judge model to extract the objects and behaviors in the description, then calculates F1 score in terms of objects and behaviors respectively. Finally, we use the average F1 score to measure the performance.

\noindent\textit{\textbf{Example.}}

\begin{tcolorbox}
    \small 
    \includegraphics[width=\linewidth]{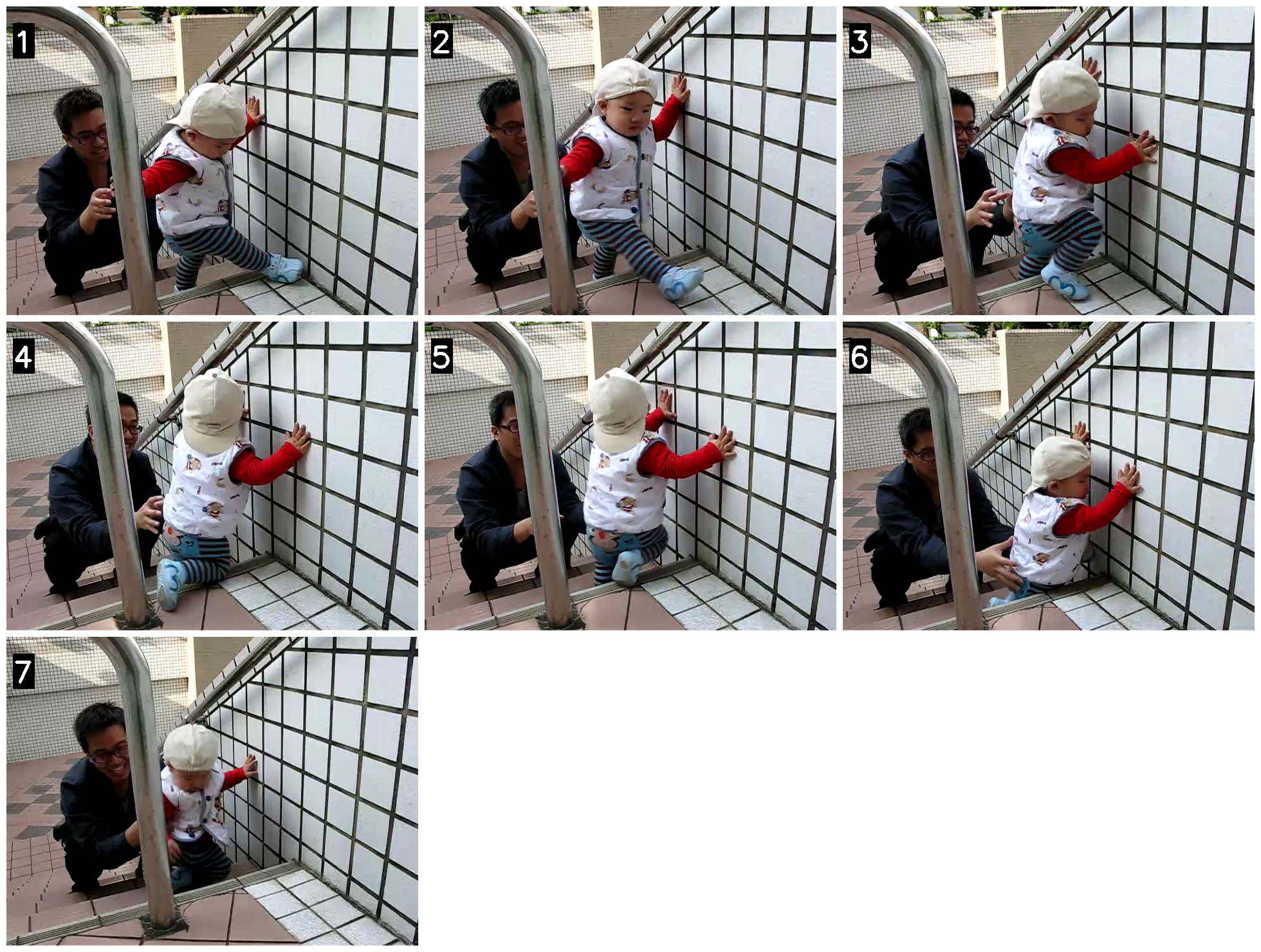} 

    \tcblower
    \small

    Write a description for the given image sequence in a single paragraph, what is happening in this episode?
    
\end{tcolorbox}

\subsection{Complementary Cases}
\label{appendix:cases}

% This section complements more cases on our evaluated benchmarks.

% \noindent\textbf{Compositional Reasoning.}

\definecolor{darkgreen}{rgb}{0.0, 0.5, 0.0}
\begin{figure}[ht] % The placement specifier [ht] can be adjusted as needed
\centering
\begin{tcolorbox}
    \small 
    \includegraphics[width=\linewidth]{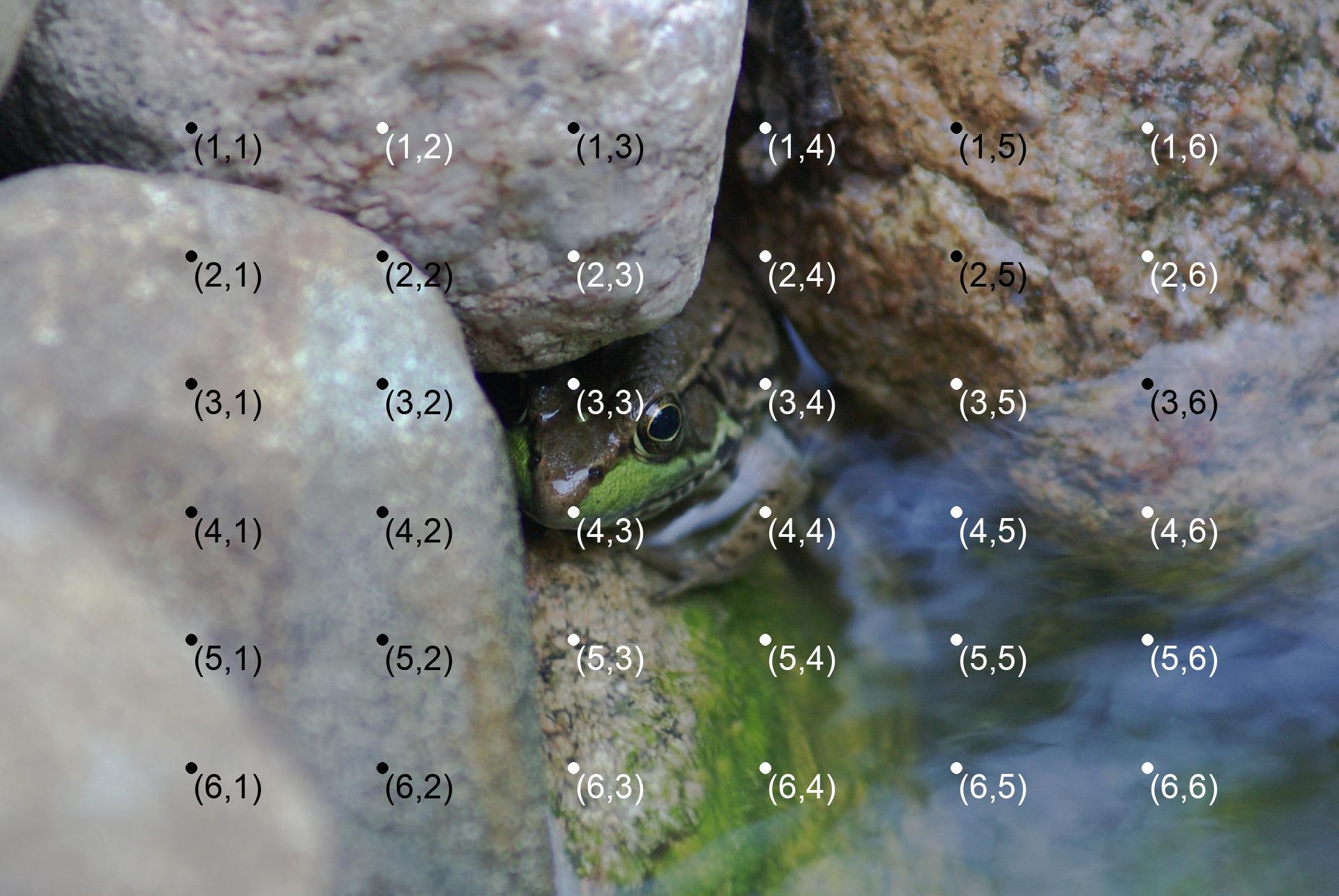} 
    \small
    \textbf{\textit{Choose the correct caption for the image. (A) a frog is above rocks which are under the water (B) a frog is under rocks which are above the water.}}

    \tcblower

    \small
    \textbf{Ground-Truth: } \textcolor{darkgreen}{B}

    \textbf{GPT-4V + CoT: } The frog is \textcolor{red}{visible above the rocks}, and the rocks are partially submerged in water, indicating that the frog is not under the rocks \textcolor{red}{but rather resting on top of them} or between them above the water level. \textcolor{red}{A}.

    \textbf{GPT-4V + \method: } \textcolor{darkgreen}{The frog is located near coordinates (3,3) and (4,3)}, and it is clearly visible underneath \textcolor{darkgreen}{the rocks, which are near coordinates (1,3), (2,3), (1,4), and (2,4)}. \textcolor{darkgreen}{The water is visible around the frog, particularly near coordinates (5,4) and (6,4)}. Based on the spatial relationships, the rocks are above the water, and the frog is under the rocks, which are above the water. The correct caption for the image is: \textcolor{darkgreen}{B}.
\end{tcolorbox}
\caption{An example from the Winoground dataset.} 
\label{fig:wino_example}
\end{figure}

\begin{figure}[ht] % The placement specifier [ht] can be adjusted as needed
\centering
\begin{tcolorbox}
    \small 
    \textit{Task: spot 10 differences in two images.}
    
    \textbf{GPT-4V: }
    
    \includegraphics[width=\linewidth]{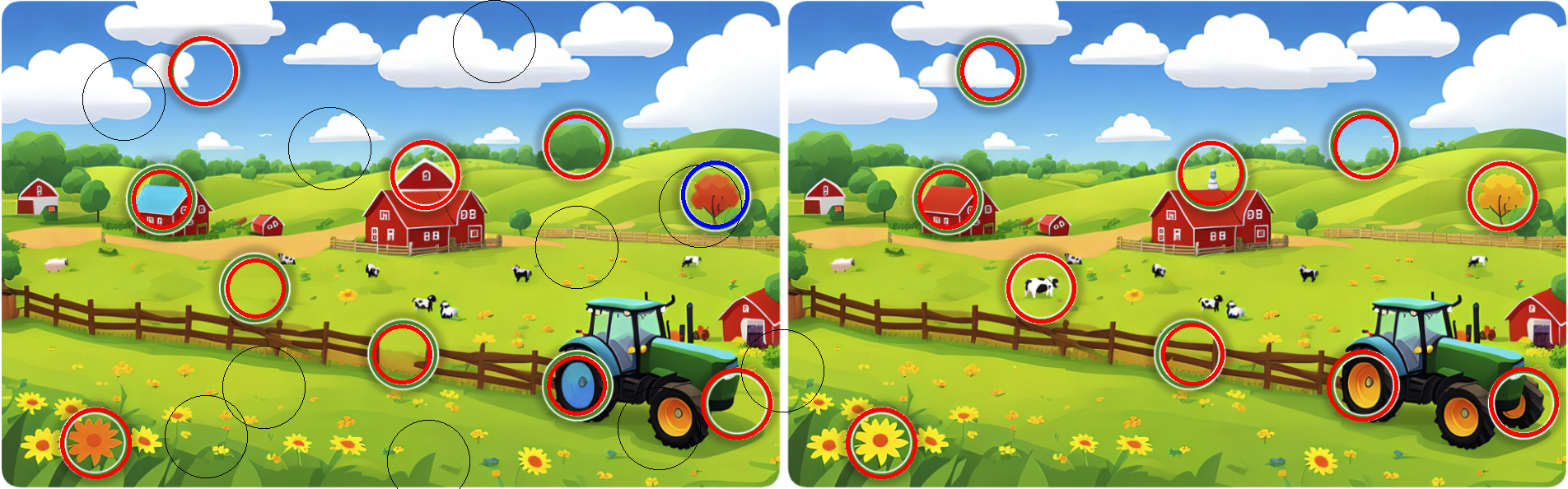} 

    \textbf{GPT-4V + \method: }
    
    \includegraphics[width=\linewidth]{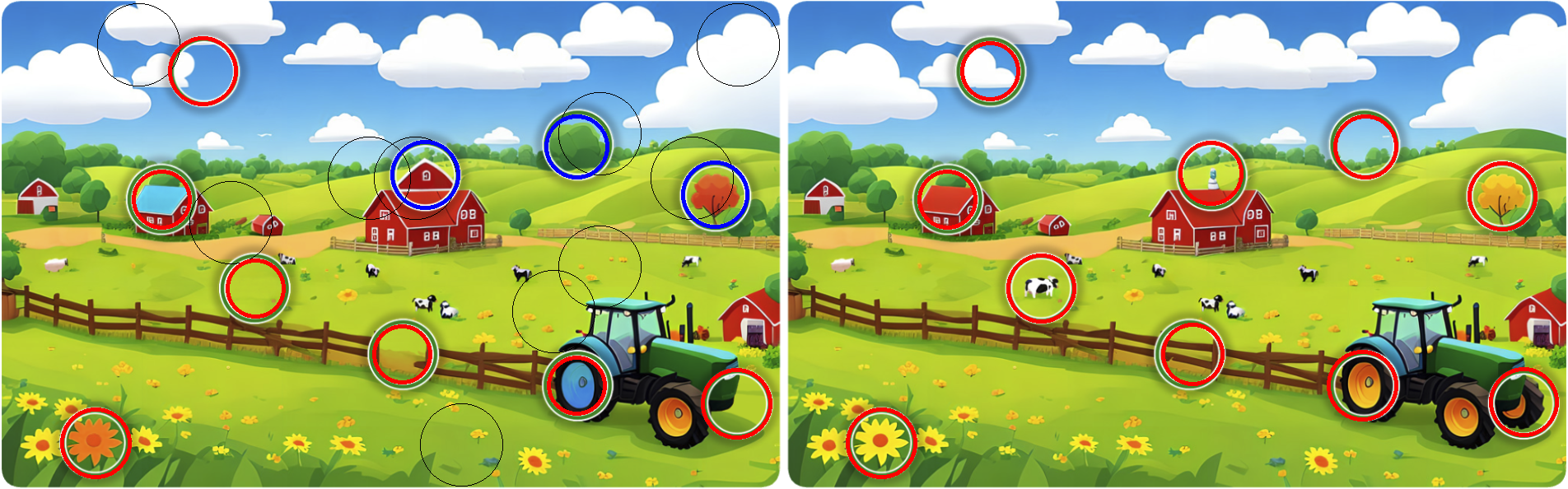} 

    \tcblower
    \small
Analysis: In this example, the addition of a dot matrix resulted in identifying a greater number of differences, reflecting the guiding role of the dot matrix in image analysis.
    
\end{tcolorbox}
\vspace{-2mm}
\caption{An example from the Spotting Differences dataset.} 
\label{fig:spotting_differences_example}
\vspace{-5mm}
\end{figure}

% \noindent\textbf{Hallucination. }

\definecolor{darkgreen}{rgb}{0.0, 0.5, 0.0}
\begin{figure}[ht] % The placement specifier [ht] can be adjusted as needed
\centering
\begin{tcolorbox}
    \small 
    \includegraphics[width=\linewidth]{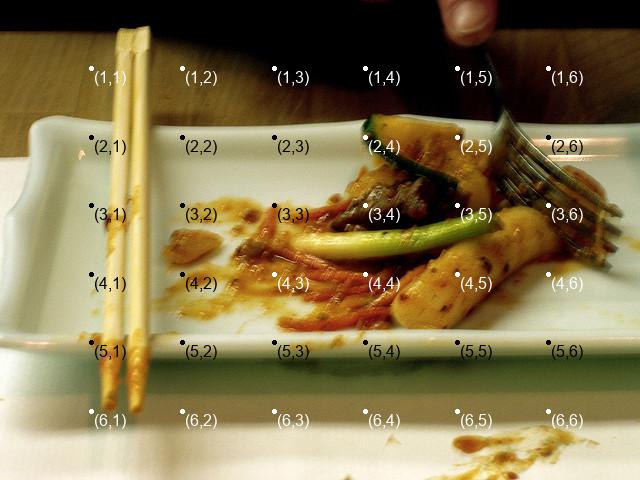} 
    \small
    \textbf{\textit{Question: Is there a dining table in the image?}}

    \tcblower
    \small
    
    \textbf{Ground-Truth: } \textcolor{darkgreen}{Yes.}

    \textbf{GPT-4V + CoT: } There is \textcolor{red}{no person visible in the image}. The image shows a plate with some food and a pair of chopsticks, and there is a fork in the process of picking up food, but \textcolor{red}{no person is seen in the frame}. So, the answer is \textcolor{red}{no.}.

    \textbf{GPT-4V + \method: } In the provided images, there is no complete person visible. However, \textcolor{darkgreen}{there is a hand holding a fork in the upper right area of the image, near coordinates (1,6) through (3,6)}. This indicates the presence of a person, but only a part of them (the hand) is shown. Therefore, the answer is \textcolor{darkgreen}{yes}.
\end{tcolorbox}
\vspace{-2mm}
\caption{An example from the POPE (adversarial) dataset.} 
\label{fig:pope_example}
\vspace{-8mm}
\end{figure}

\begin{figure*}[h]
    \centering
    % First Row
    \begin{subfigure}{.5\textwidth}
        \centering
        \includegraphics[width=.9\linewidth]{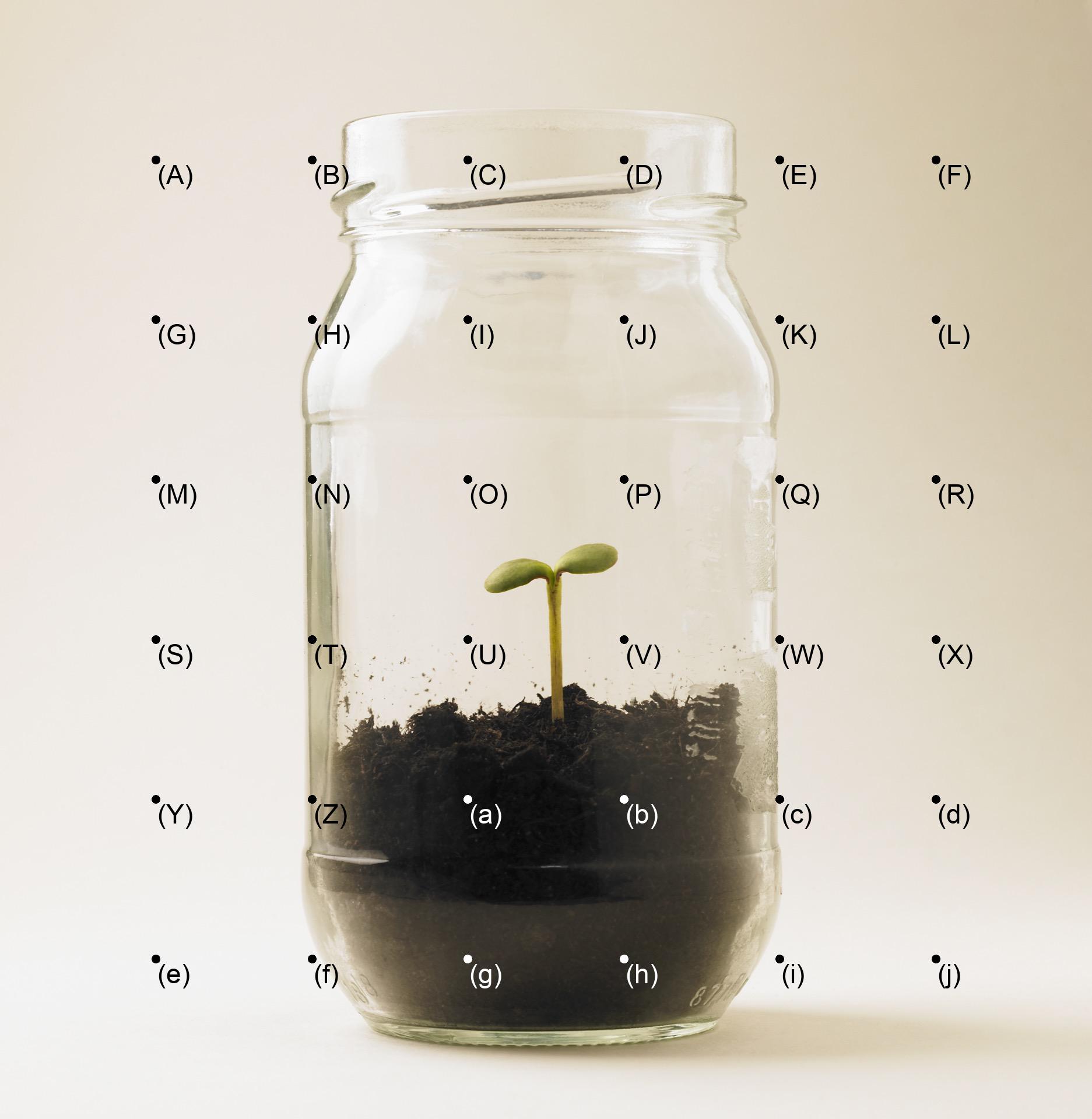}
        \caption{Alphabet}
    \end{subfigure}%
    \begin{subfigure}{.5\textwidth}
        \centering
        \includegraphics[width=.9\linewidth]{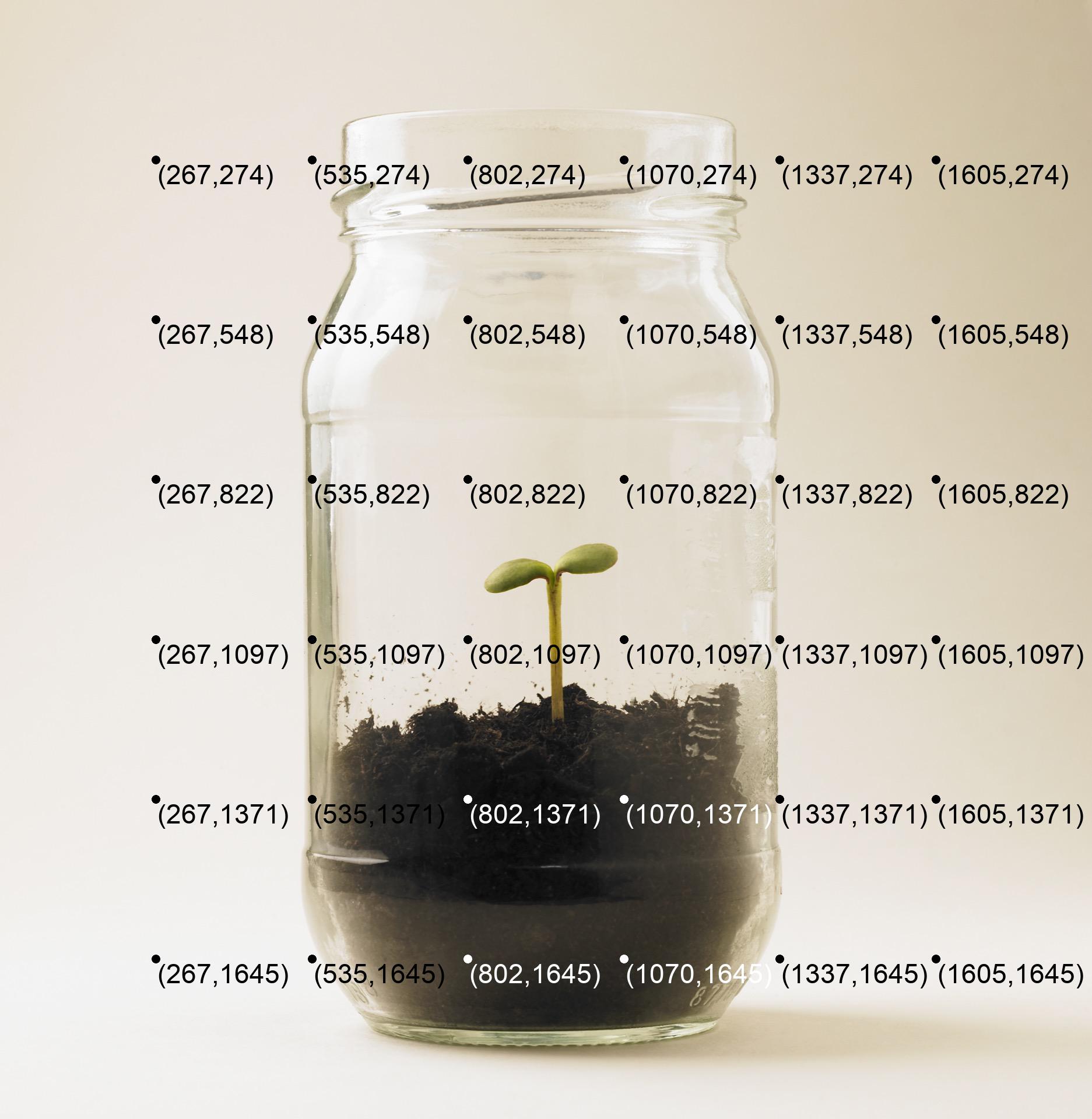}
        \caption{Pixel}
    \end{subfigure}
    
    % Second Row
    \begin{subfigure}{.5\textwidth}
        \centering
        \includegraphics[width=.9\linewidth]{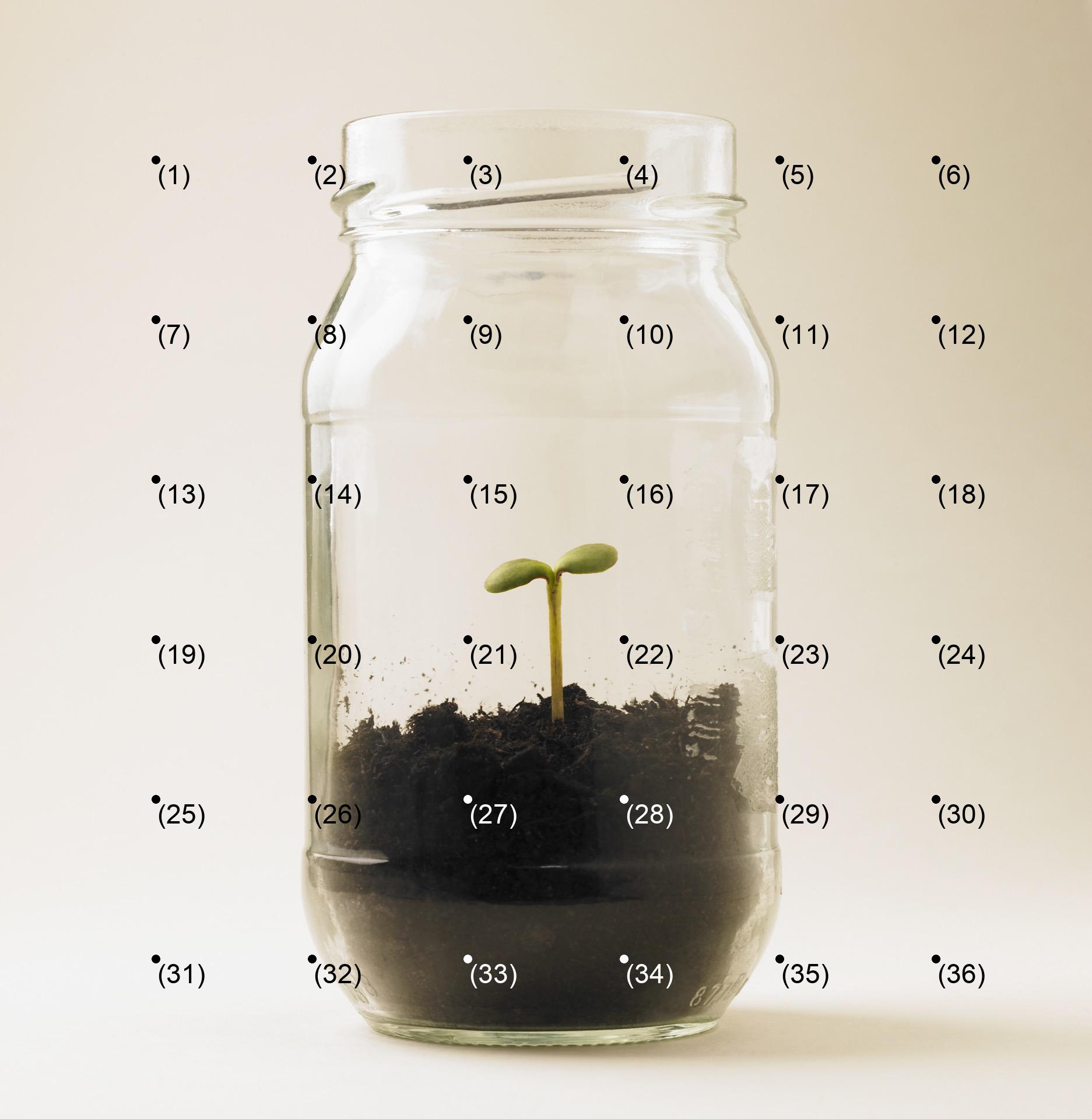}
        \caption{One-dimensional}
    \end{subfigure}%
    \begin{subfigure}{.5\textwidth}
        \centering
        \includegraphics[width=.9\linewidth]{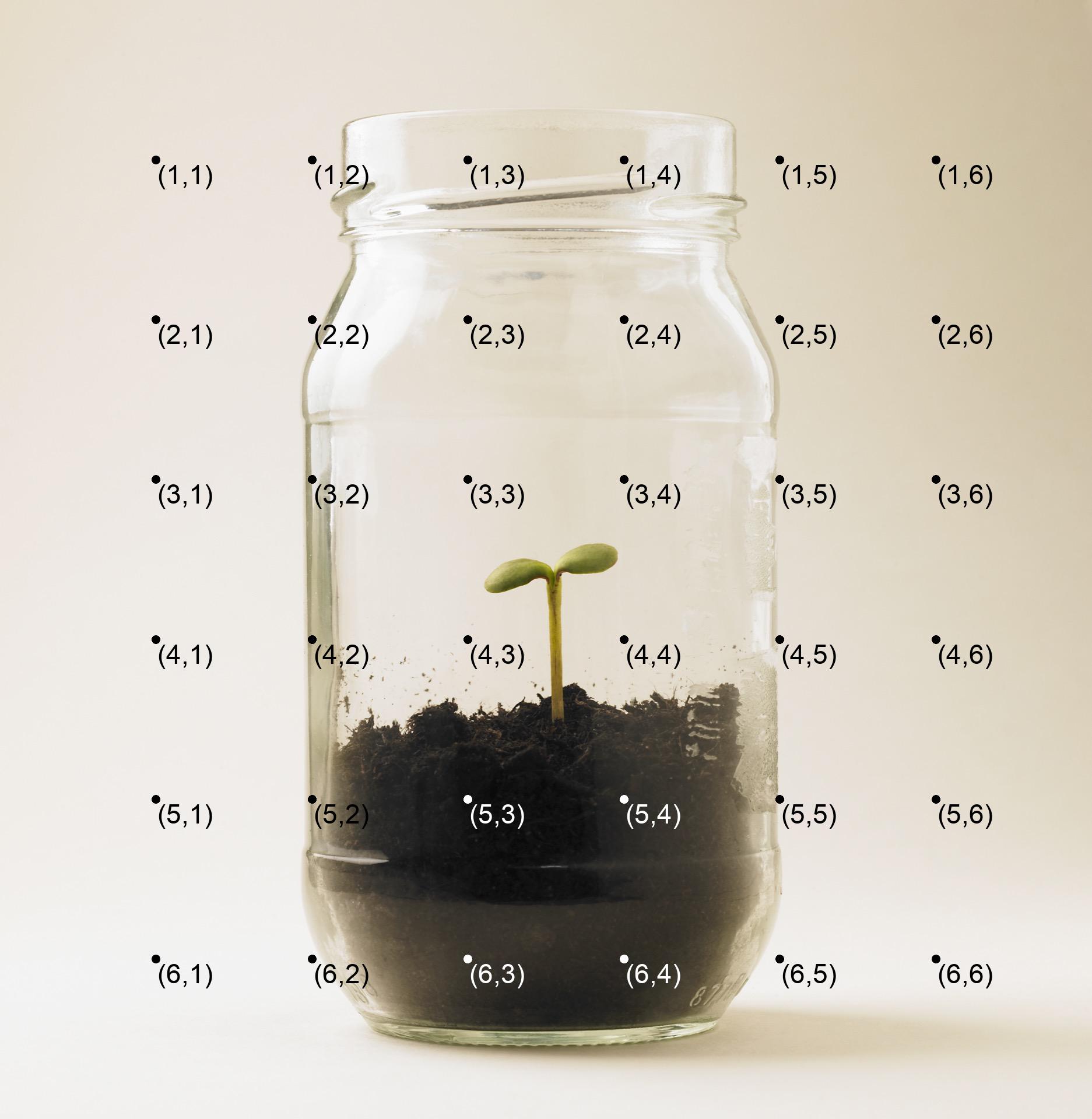}
        \caption{Cartesian (Ours)}
    \end{subfigure}
    
    \caption{Examples of different coordinate formats.}
    \label{fig:format_comparison}
\end{figure*}

\end{document}